\documentclass[10pt,conference]{IEEEtran}
\IEEEoverridecommandlockouts
\usepackage{cite}
\newtheorem{theorem}{Theorem}[section]

\newtheorem{definition}{Definition}[section]

\usepackage{forest}
\usepackage{amsmath,amssymb,amsfonts}
\usepackage{algorithmic}
\usepackage{graphicx}
\usepackage{textcomp}
\usepackage{xcolor}
\usepackage{tikz}
\usepackage{multirow}
\usepackage{hyperref}
\def\BibTeX{{\rm B\kern-.05em{\sc i\kern-.025em b}\kern-.08em
    T\kern-.1667em\lower.7ex\hbox{E}\kern-.125emX}}
\makeatletter
\def\ps@IEEEtitlepagestyle{%
	\def\@oddfoot{\mycopyrightnotice}%
	\def\@evenfoot{}%
}
\def\mycopyrightnotice{%
	{\footnotesize manuscript is under review \hfill}
	\gdef\mycopyrightnotice{}
}
\begin{document}

\title{Over-Squashing in Graph Neural Networks:\\ A Comprehensive survey\\
	{
	}
}

\author{\IEEEauthorblockN{S Akansha}
	\IEEEauthorblockA{\textit{Department of Mathematics}\\
		\textit{Manipal Institute of Technology}\\
		Manipal Academy of Higher Education - 576104, India.\\
		akansha.agrawal@manipal.edu.}
}

\maketitle

\begin{abstract}
Graph Neural Networks (GNNs) revolutionize machine learning for graph-structured data, effectively capturing complex relationships. They disseminate information through interconnected nodes, but long-range interactions face challenges known as "over-squashing". This survey delves into the challenge of over-squashing in GNNs, where long-range information dissemination is hindered, impacting tasks reliant on intricate long-distance interactions. It comprehensively explores the causes, consequences, and mitigation strategies for over-squashing. Various methodologies are reviewed, including graph rewiring, novel normalization, spectral analysis, and curvature-based strategies, with a focus on their trade-offs and effectiveness. The survey also discusses the interplay between over-squashing and other GNN limitations, such as over-smoothing, and provides a taxonomy of models designed to address these issues in node and graph-level tasks. Benchmark datasets for performance evaluation are also detailed, making this survey a valuable resource for researchers and practitioners in the GNN field. 
\end{abstract}

\begin{IEEEkeywords}
Graph Neural Networks (GNNs), Over-squashing, Over-smoothing, Graph-rewiring, Graph transformers
\end{IEEEkeywords}

\section{Introduction}
In recent years, the explosion of data in various domains has led to an increased interest in harnessing the power of graph structures for modeling complex relationships \cite{ran_she_kout-15a, les_mca-12a, def_bre_van-16a, gil_sch_ril-17a, ham_yin_les-17a}. Graphs, which consist of nodes and edges representing entities and their connections, respectively, have emerged as a fundamental data representation in fields such as social networks \cite{les_mca-12a,che_li_bru-17a,min_gao_pen-21a}, recommendation systems\cite{wan_yuy-22a,gao_wan-22a,chu_yao-22a,che_yeh_wan-22a}, biology \cite{yan_li-23a,jin_eis_son-21a}, and more. As the diversity and complexity of graph-structured data grow, so does the demand for advanced tools to analyze and understand these intricate relationships.

The burgeoning interest in leveraging graph-structured data has given rise to a remarkable class of machine learning models known as Graph Neural Networks (GNNs) \cite{yua_yu_gui-22a, zho_cui_hu-20a, wu_pan_che-20a}. GNNs represent a novel approach to learning representations, allowing the unified capture of both local and global information of nodes within graphs \cite{xu_hu_les-18a, sat-20a}. In essence, GNNs extend traditional neural network architectures to accommodate graph data, where nodes symbolize entities and edges signify relationships. This extension has paved the way for a myriad of applications, spanning from node classification \cite{xia_wan_dai-22a, par_nev-2019a, wan_jin_zha-21a} and link prediction to broader graph-level tasks such as community detection \cite{che_li_bru-17a, qiu_hua_xu-22a} and molecular property prediction \cite{wie_koh-20a, wan_liu_luo-22a}. By leveraging the underlying graph structure, GNNs facilitate information propagation and aggregation, allowing them to capture intricate patterns that traditional machine learning models find challenging to discern.

Notwithstanding their remarkable achievements, GNNs are not immune to certain inherent limitations, including over-smoothing \cite{wu_che_wan-22a, rus_bro_mis-23a}, vanishing gradients \cite{luk_leh_fis-20a, wu_jai_wri-21a}, Out-of-Distribution (OOD) data challenges \cite{mah_swe_kip-22a, aka-2023a}, overfitting \cite{bo_hu_wan-22a}, and the relatively less explored phenomenon of over-squashing \cite{alo_yah-20a, gir_mal_bou-22a, gio_giu_far-23a}. While exhaustive research has been dedicated to addressing the former issues, the latter—over-squashing—remains relatively less explored.

Over-squashing is a phenomenon that manifests in tasks requiring the integration of information from distant nodes\cite{alo_yah-20a, top_gio_cha-21a}, primarily through edges that serve as bottlenecks within graph data. To put it succinctly, over-squashing denotes the distortion-prone nature of information transfer between nodes that are widely separated\cite{gio_giu_far-23a}. This distortion emerges due to the inherent tension between the limited feature representation capacity of graph embeddings and the exponential growth in the number of neighbors as graphs expand. This interplay often hampers the faithful transmission of distant information.

This survey article aims to provide a comprehensive panorama of this specific limitation. We delve into the intricate nuances of over-squashing, shedding light on its conceptual framework and its implications. Additionally, we meticulously outline the repertoire of methods proposed thus far to grapple with this intricate issue. By presenting a systematic exploration of the landscape, we contribute to a deeper understanding of over-squashing's impact on GNNs and offer insights into the evolving strategies engineered to surmount this challenge.

To summarize, this paper makes the following key contributions:
\begin{enumerate}
\item \textit{Pioneering Survey:} This paper serves as the inaugural comprehensive survey on 'over-squashing,' a pivotal limitation in message-passing graph neural networks. It addresses a burgeoning area of interest among researchers.
\item \textit{Systematic Categorization:} We provide a systematic categorization of existing methods, offering a detailed taxonomy that simplifies the understanding of various strategies to mitigate over-squashing.
\item \textit{Benchmark Datasets:} We extensively discuss commonly used benchmark datasets employed for evaluating models in the context of over-squashing, both at the node and graph levels.
\item \textit{Added Value:} Additionally, this survey explores the interplay of over-squashing with other fundamental GNN limitations, such as 'over-smoothing,' providing a more holistic perspective on the challenges faced in this domain.
\end{enumerate}
\textbf{Methodology for Literature Review: }To ensure a comprehensive and structured review, we followed a systematic methodology to identify relevant literature on \textbf{over-squashing in Graph Neural Networks (GNNs)} and related mitigation techniques. The key steps in our methodology are as follows:
\begin{enumerate}
	\item \textbf{Database Selection:} 
	We primarily utilized \textit{Google Scholar} for our search, given its extensive coverage of academic publications across conferences, journals, and preprint servers.
	
	\item \textbf{Search Strategy:} 
	Our search process began by identifying key papers published in top-tier machine learning conferences such as \textit{ICML}, \textit{ICLR}, and \textit{NeurIPS}, which are known for hosting pioneering research on GNNs and related topics.
	
	\item \textbf{Keyword Selection:} 
	To expand the search, we used the following keywords in combination:``over-squashing in GNNs'', ``bottleneck effect in GNNs'', ``rewiring techniques in GNNs''.
	
	\item \textbf{Citation Tracking:} 
	After identifying foundational papers, we examined their \textit{citation networks} to discover additional relevant works that extended or improved upon the original methodologies. This backward and forward citation tracking allowed us to include both seminal contributions and recent advancements in the field.
	
	\item \textbf{Inclusion Criteria:} 
	Papers were selected based on their relevance to over-squashing issues in GNNs, proposed solutions such as graph rewiring, and their impact as indicated by citations and peer recognition.
\end{enumerate}

Our systematic approach ensured that the survey encompasses both well-established techniques and emerging methods designed to address over-squashing challenges in GNNs. While there are existing survey articles in the literature that explore specific aspects of this problem, they focus on narrower themes. For instance, the survey by Attali et al. \cite{att-bus_24a} specifically discusses curvature-based rewiring techniques for mitigating over-squashing in GNNs, while Yadati \cite{yad_24a} examine over-squashing within the context of hypergraph neural networks. In contrast, our survey provides a more comprehensive overview by consolidating a broader range of methods, including spatial, spectral, and other approaches. To the best of our knowledge, this makes our survey the most extensive resource that offers a holistic perspective on mitigating over-squashing in GNNs.


This article is organized as follows: In Section \ref{sec:background}, we provide a comprehensive background on graph neural networks (GNNs) (\ref{subsec:gnn}), detailing the concept of over-squashing and its mathematical underpinnings. Section \ref{sec:handlingosq} delves into popular methods for mitigating over-squashing in GNNs, including spatial graph rewiring (\ref{subsec:spatialrewiring}) and spectral graph rewiring (\ref{subsec:spectralrewiring}). In Section \ref{sec:osqosm}, we explore the intricate interplay between over-squashing and over-smoothing in GNNs, analyzing their trade-offs and presenting unified approaches to address both challenges simultaneously. Section \ref{sec:otherarchitecture} examines novel techniques beyond rewiring methods that tackle over-squashing. Furthermore, Section \ref{sec:datasets} introduces commonly used benchmark datasets for evaluating over-squashing in GNNs, applicable to both node classification and graph classification tasks. Finally, Section \ref{sec:conclusion} offers concluding remarks and outlines promising avenues for future research, providing insights into underexplored directions for further exploration.
\section{Background} \label{sec:background}
\subsection{Graph Neural Networks}\label{subsec:gnn}

A Graph Neural Network is a specialized neural network architecture designed for processing data structured as graphs. The central concept behind GNNs involves the iterative collection of information from neighboring nodes and the updating of node features through multiple layers.

Let's delve deeper into this concept. Imagine a graph, denoted as $G = (V, E)$, where $V$ represents the set of nodes with $|V|=n$ and $E$ signifies the set of edges (or links), adhering to $E \subseteq V \times V$. If for \(u,v\in V\) is connected by an edge, we write it as \((u,v)\in E\). The adjacency matrix $A$ is defined by $A_{uv} = 1$ if $(u,v) \in E$, and $A_{ij} = 0$ otherwise. Within this framework, each node $v\in V$ comes with an associated feature vector $x_v\in \mathbb{R}^{d_0}$ effectively encapsulating information pertaining to the attributes of the node $v$. We use \(X\in \mathbb{R}^{n\times d_0}\) as feature matrix. In the realm of GNNs, the primary goal is to acquire effective representations for nodes, links, and even entire graphs. This is achieved through a fundamental process known as message-passing, initially defined by Gilmer et al. \cite{gil_sch_ril-17a} and further elaborated by Zhang et al. \cite{zha_yao-22a}.

In this message-passing process, GNNs iteratively enhance node representations using the following equations:

At layer $l$:
\begin{align}\label{eq:gnnlayer}
	h^{(l)}_u & = UP_{l}\biggl\{h^{(l-1)}_u, \vspace{1cm} AGG_{l}\{h^{(l-1)}_v \mbox{ where } v \in N_u\}\biggr\}
\end{align}
Here, $h^{(l-1)}_u$ represents the node representation at the $(l-1)$-st layer, typically initialized with the node's feature at the initial layer $h^{(0)}_u = x_u$. $N_u$ denotes the set of $1$-hop neighboring nodes of node $u$:
\[N_u = \{v\in V : (v,u)\in E\}.\]  The aggregation function family, denoted as $AGG_{l}(\cdot)$, is responsible for mixing the information from neighboring nodes and has the form: 
$$AGG_{l} : \mathbb{R}^{d_{(l-1)}} \times \mathbb{R}^{d_{(l-1)}} \to \mathbb{R}^{d_{(l-1)}'}$$
Furthermore, the update function family, referred to as $UP_{l}(\cdot)$, integrates the aggregated information into the node's representation and has the form: 
$$UP_{l} : \mathbb{R}^{d_{(l-1)}} \times \mathbb{R}^{d_{(l-1)}'} \to \mathbb{R}^{d_{(l)}}$$
In the context of message passing GNNs, $AGG$ and $UP$ typically serve as attention mechanisms and activation functions, respectively. Through iterative application of this message-passing mechanism, GNNs progressively enhance node representations by accounting for their associations with neighboring nodes. This iterative refinement process plays a vital role in capturing both structural and semantic insights embedded within the graph.



\subsection{Over-squashing}\label{subsec:osqanalysis} 
The issue of over-squashing is a recognized challenge encountered in Message Passing Neural Networks (MPNNs) when messages traverse distant nodes. This problem arises due to the rapid expansion of a node's receptive field, leading to the compression of numerous messages into fixed-size vectors. Topping et al. \cite{top_gio_cha-21a} have formally substantiated this phenomenon through a sensitivity analysis of the Jacobian matrix of node features. 


To elaborate further, let's consider a receptive field denoted as \( B_r = \{ v \in V : d_G (u, v) \leq r \} \) associated with an \( r \)-layer GNN. Here, \( d_G \) represents the shortest-path distance between node \( u \) and node \( v \), and \( r \) is a natural number. The shortest-path between node \(u\) and \(v\) is denoted by \(P_{uv}\). The Jacobian matrix \( \frac{\partial h^{(r)}_u}{\partial x_v} \) quantifies the sensitivity of a node representation \( h^{(r)}_u \) to a specific input feature \( x_v \) in node \( v \). Over-squashing can be conceptualized as the incapacity of \( h^{(r)}_u \) to be influenced by \( x_v \) at a distance of \( r \). Topping et al. \cite{top_gio_cha-21a} have established the following bounds to quantify the effect of over-squashing:

\begin{equation}\label{eq:osq_bound_top}
\frac{\partial h^{(r+1)}_u}{\partial x_v} \leq (\alpha \beta)^{r+1}(\tilde{A}^{r+1})_{uv}.
\end{equation}
In equation \eqref{eq:osq_bound_top}, $\tilde{A}= D^{-\frac{1}{2}}(A+I)D^{-\frac{1}{2}}$ represents the normalized adjacency matrix, where $D_{ii}= \sum_{j=1}^{n}A_{ij}$ stands for the degree matrix of $(A+I)$. Note that ${A}_{ij}$ denotes the entry in row $i$ and column $j$ of matrix ${A}$. Inequality \eqref{eq:osq_bound_top} conveys that if $|\nabla UP_l| \leq \alpha$ and $|\nabla AGG_l| \leq \beta$ for $0 \leq l \leq r+1$, where $\nabla g$ denotes the Jacobian of a map $g$, the propagation of messages can be controlled by an appropriate power of $\tilde{A}$. Specifically, this inequality emphasizes how the influence of input features diminishes exponentially as the distance $r$ increases, which becomes particularly pronounced when the size of $B_r$ grows exponentially.

On the same assumption as for bound \eqref{eq:osq_bound_top}, Black et al. \cite{bla_wan_nay-23a} provide the following bound:
\begin{equation}\label{eq:osq_bound_bla}
\frac{\partial h^{(r+1)}_u}{\partial x_v} \leq (2 \alpha \beta) \sum_{l=0}^{r+1} (\tilde{A}^{l})_{uv}.
\end{equation}
The primary distinction between bounds \eqref{eq:osq_bound_top} and \eqref{eq:osq_bound_bla} lies in their conditions: the former holds when the vertices $u$ and $v$ must be precisely at a distance $r+1$ from each other, while the latter applies to any pair of vertices.
 

Black et al. \cite{bla_wan_nay-23a} additionally established a bound to quantify the over-squashing effect, which correlates heightened effective resistance between node pairs to the sensitivity of a node representation \( h^{(r)}_u \) to a specific input feature \( x_v \). 
\begin{definition}
	The effective resistance between two nodes \( u \) and \( v \) is defined as:
	\begin{equation}\label{eq:eff_resistance}
		R_{u,v} = \left( \frac{1}{\sqrt{d_u}} \mathbf{1}_u - \frac{1}{\sqrt{d_v}} \mathbf{1}_v \right)^T \tilde{L}^+ \left( \frac{1}{\sqrt{d_u}} \mathbf{1}_u - \frac{1}{\sqrt{d_v}} \mathbf{1}_v \right),
	\end{equation}
	where \( \mathbf{1}_v \) is the indicator vector of vertex \( v \), \( d_v \) denotes the degree of vertex \( v \), and \( \tilde{L}^+ \) is the pseudoinverse of the normalized Laplacian matrix \( \tilde{L} = I - \tilde{A} \). 
\end{definition}

The effective resistance quantifies the level of connectivity between vertices \( u \) and \( v \) within the graph \( G \). Essentially, when multiple paths exist between two nodes, the effective resistance \( R_{u,v} \) tends to be small, signifying higher connectivity. Conversely, when the available paths between two nodes are limited, \( R_{u,v} \) becomes larger, indicating lower connectivity.

Based on the effective resistance between nodes $u$ and $v$, Black et al \cite{bla_wan_nay-23a} provide the following bound to quantify the impact of each node representation $h^{(r)}_u$ with respect to $x_v$. 
\begin{theorem}\label{thm:osq_bound_effresis_bla} For a connected graph \( G \), let \( u, v \in V \) and \( \| \nabla UP_l \| \leq \alpha \) and \( \max \{ \| \nabla AGG_l \|, 1 \} \leq \beta \) for all \( l = 0, \ldots, r \). Let \( d_{\text{max}} \) and \( d_{\text{min}} \) be the maximum and minimum degrees of nodes \( u \) and \( v \), and \( \max \{|\mu_2|, |\mu_n|\} \leq \mu \) where \( \mu_n \leq \mu_{n-1} \leq \cdots \leq \mu_1 = 1 \) denote the eigenvalues of \( \tilde{A} \). Then, 
\begin{equation}\label{eq:osq_bound_effresis_bla}
	\left\| \frac{\partial h^{(r)}_u}{\partial x_v} \right\| \leq (2\alpha\beta)^r \frac{d_{\text{max}}}{2} \left(\frac{2}{d_{\text{min}}}\left(r+1+\frac{\mu^{r+1}}{1-\mu}\right) - R_{u,v} \right).
\end{equation}
\end{theorem}
Theorem \ref{thm:osq_bound_effresis_bla} implies that vertices with lower effective resistance exert a stronger influence on each other during message passing. Specifically, the node feature \( h^{(r)}_u \) at node \( u \) in layer \( r \) is more heavily influenced by the initial node feature \( x_v \) at node \( v \). Hence, it characterizes the over-squashing effect in GNN as proportional to the effective resistance between nodes of the graph. This notion aligns with intuition, as effective resistance reflects the connectivity and number of paths between \( u \) and \( v \). When there are numerous short paths connecting \( u \) and \( v \), the effective resistance between them decreases, facilitating stronger interaction hence low over-squashing. In a parallel vein, Di Giovanni et al. \cite{gio_giu_far-23a} have undertaken similar methodologies, ultimately converging on a shared conclusion. Their findings underline the pivotal role of effective resistance in influencing the degree of over-squashing within GNNs. 
They delve into the impact of GNN architecture's width and depth on the occurrence of over-squashing. In their work, Di Giovanni et al. \cite{gio_giu_far-23a} build upon their previous findings and concentrate on two pivotal factors: the network's architecture, characterized by weight norms and depth, and the intrinsic graph structure, evaluated using commute times. Commuting time, in the context of a graph, refers to the expected number of steps it takes for a random walk to travel between two specific nodes. It's like measuring how long it would typically take for someone to move from one node to another and coming back by randomly traversing the edges of the graph. It is well known that \cite{cha-rag_89a} commute time between node $u$ and $v$ (denoted as $Com(u, v)$) is directly proportional to effective resistance $R_{u,v}$. In mathematical terms, this relationship is expressed as $Com(u, v) = 2|E|R_{u,v}$, with $E$ representing the set of edges in the graph.

In \cite{gio_giu_far-23a}, the quantification of the influence of \(x_v\) on the node representation of node \(u\) at any layer \(l<r\) is refined as the symmetric Jacobian obstruction between nodes \(u\) and \(v\). This refinement is achieved by defining 
\begin{equation}
O^{(r)}_{u,v} = \sum_{l=0}^{r} \left\|J^{(r)}_{l}(u,v)\right \|,
\end{equation}
where $\|\cdot\|$ denotes the spectral norm of a matrix and \(J^{(r)}_{l}\) is calculated using the formula:
\[ J^{(r)}_{l}(u,v) = \frac{1}{d_u} \frac{\partial h^{(r)}_u}{\partial h^{(l)}_u} - \frac{1}{\sqrt{d_u d_v}} \frac{\partial h^{(r)}_u}{\partial h^{(l)}_v} + \frac{1}{d_v} \frac{\partial h^{(r)}_v}{\partial h^{(l)}_v} - \frac{1}{\sqrt{d_u d_v}} \frac{\partial h^{(r)}_v}{\partial h^{(l)}_u} \]

Intuitively, \(J^{(r)}_{l}(u,v)\) reflects the sensitivity between nodes \(u\) and \(v\). When nodes \(u\) and \(v\) are less sensitive to each other, \(J^{(r)}_{l}(u,v)\) tends to be larger; conversely, it is smaller when the communication is robust. \(O^{(r)}_{u,v}\) quantifies the influence of the initial feature of node \(v\) on \(h^{(r)}_u\) and can be bounded by the commute time.
\begin{theorem}\label{thm:osqbound_ct}
For a specific MPNN as in \eqref{eq:gnnlayer} with  \(AGG_l = W^{(l)}\). Let \( \nu \) be the minimal singular value  and \(w\) be the maximal spectral norm of the matrix \(W^{(l)}\). Assuming all paths in the MPNN graph are activated with success probability \( \rho \) then there exists a constant \( \epsilon_G \) independent of \( u \) and \( v \), such that:
\begin{equation*}
\epsilon_G (1 - o(l)) \frac{\rho}{\nu} \frac{Com(u, v)}{2|E|} \leq O^{(l)}_{u,v} \leq \frac{\rho}{w} \frac{Com(u, v)}{2|E|}
\end{equation*}
where \( o(l) \) approaches $0$ exponentially fast as \( l \) increases.
\end{theorem}
Previously discussed, a smaller value of \(O^{(r)}_{u,v}\) indicates that node \(u\) is more sensitive to \(v\) in the MPNN, and vice versa. Therefore, Theorem \ref{thm:osqbound_ct} suggests that nodes with shorter commute times will exchange information more effectively in an MPNN, while conversely, nodes with longer commute times will exchange information less effectively. Consequently, they infer that over-squashing becomes problematic when the task relies on interactions between nodes with high commute times.

In previous works \cite{top_gio_cha-21a,bla_wan_nay-23a,gio_giu_far-23a}, researchers have formalized the phenomenon of over-squashing by examining how the Jacobian of node features is influenced by topological properties of the graph, such as curvature, effective resistance, or commute time. However, in \cite{gio_rus_bro-23a}, the authors conducted a more comprehensive analysis and introduced a novel over-squashing metric. This metric takes into account various aspects including GNN architecture, parameters, graph topology, and downstream tasks. As part of their investigation, they established upper bounds on the capacity of MPNNs, defined by the pair $(r,w)$, where $r$ represents the number of layers and $w$ denotes the maximum spectral norm of the weights. A higher capacity, achieved by increasing either $r$ or $w$, or both, indicates greater expressive capability of the MPNN, allowing for more extensive mixing among node features. Importantly, they introduced the concept of "over-squashing," which is intricately linked to the maximum node mixing capacity of MPNNs and operates inversely to it.

\begin{definition}\label{def:max_mixing}
Consider a (twice differentiable) function \( y_G \) (in the context of MPNN it serves as ground truth function for graph-level tasks). The maximal mixing induced by \( y_G \) among variables \( x_u \) and \( x_v \), where \( u \neq v \), is given by:
\[
\text{mix}_{y_G}(u, v) = \max_{x_i} \max_{1 \leq \alpha, \beta \leq d_0} \left\vert \frac{\partial^2 y_G(X)}{\partial x_{\alpha}(u) \partial x_{\beta}(v)} \right\vert,
\]
\end{definition}
where \(x_u(i)\) denotes the \(i\)th component in the vector \(x_u\).

\begin{definition}
In the context of an MPNN comprising \( r \) layers and \( w \) representing the maximum spectral norm of its weights, we define the tuple \((r, w)\) as indicative of the MPNN's capacity.
\end{definition}
The over-squashing of nodes \(v\) and \(u\) in an MPNN, characterized by its capacity \((m,w)\), is quantified by the function \(OSQ_{v,u}(m,w)\). For a given MPNN with capacity \((r,w)\), the over-squashing effect between nodes \(u\) and \(v\) can be characterized by the pairwise mixing induced by an MPNN with the same capacity \((r,w)\). Specifically, higher over-squashing between \(u\) and \(v\) corresponds to lower mixing. This relationship can be expressed as:

\[
OSQ_{u,v}(r,w) < \left( \text{mix}_{y_G}(u, v) \right)^{-1}
\]

Hence, an MPNN operating on a graph \( G \) with node features \( X \) may exhibit over-squashing depending on the complexity of the task it's designed to solve. This phenomenon varies based on the nodes \( v \) and \( u \) involved, being more pronounced for nodes with greater separation or higher commute time, highlighting the significant influence of graph topology on over-squashing. 

\section{Handling over-squashing in GNNs}\label{sec:handlingosq}
In scenarios where tasks necessitate spanning multiple layers within a network, the depth of the network often mirrors the range of interactions between nodes. Nevertheless, a rising number of layers corresponds to an exponential increase in the number of nodes contributing to the receptive field of each individual node. This amplification leads to the phenomenon of over-squashing \cite{alo_yah-20a, top_gio_cha-21a}. Essentially, over-squashing manifests as a compression of information originating from a receptive field that encompasses numerous nodes. This compression results in fixed-length node vectors, impeding the accurate propagation of messages from distant nodes. This distortion takes shape due to graph bottlenecks that emerge as the number of $k$-hop neighbors undergoes exponential growth with each $k$.

In a bid to surmount these challenges, the literature has proposed strategies such as graph rewiring \cite{alo_yah-20a,top_gio_cha-21a,ngu_hie-23a} and pooling\cite{yin_you_mor-18a, luz_day_lio-19a, san_rot_lie-23a}. \textbf{Graph rewiring} is a technique used to modify the structure of a graph, particularly the edges, with the goal of enhancing the performance of GNNs. Essentially, rewiring entails applying an operation \( R \) to \( G = (V, E) \), resulting in a new graph \( R(G) = (V, R(E)) \) with modified connectivity while retaining the same set of vertices. We extend the MPNN framework \eqref{eq:gnnlayer} to incorporate the rewiring operation \( R \) as follows:
\begin{multline}\label{eq:rewired_gnnlayer}
h^{(l)}_u = \text{UP}_{l} \left( h^{(l-1)}_u, {AGG_{(l)}}_G\left(\{ h^{(l-1)}_v : (u, v) \in E \}\right),\right.\\ \left.{AGG_{(l)}}_{R(G)}\left(\{ h^{(l-1)}_v : (u, v) \in R(E) \}\right) \right),
\end{multline}

where node features are updated based on the information gathered from both the original graph \( G \) and the rewired graph \( R(G) \) using potentially independent aggregation maps. Many GNN models based on rewiring simply exchange messages over \( R(G) \), i.e., they utilize \( {AGG_{(l)}}_G = 0 \). 

Graph rewiring typically involves making adjustments to the edges, either by adding new edges \cite{kar_ban-22a, bla_wan_nay-23a}, removing existing ones \cite{liu_zho_pan-23a}, a combination of both\cite{gir_mal_bou-22a} or replacing existing ones by new ones \cite{top_gio_cha-21a, ban_kar_wan-22a}. The primary objective of graph rewiring is to optimize the flow of information within the graph, making it more suitable for specific tasks like graph or node classification, as well as link prediction. Many GNNs implicitly incorporate the concept of graph rewiring, which involves various techniques such as utilizing Cayley graphs \cite{dea_lac_vel-22a} or introducing virtual nodes \cite{cai-tru_23a}. Some research explores directly modifying the graph connectivity to address noise \cite{gas_etal-19b} or facilitate multi-hop aggregations \cite{wan-yin_20a}.




The challenge of over-squashing within GNNs has spurred the development of various methodologies, each aiming to alleviate this phenomenon. Broadly, these methods can be categorized into two types of graph rewiring methods, each offering unique insights into the resolution of the over-squashing predicament.

\subsection{Spatial Graph Rewiring Methods:}\label{subsec:spatialrewiring} Spatial graph rewiring focuses on modifying graph connections based on geometric or spatial relationships between nodes. By optimizing these spatial connections, the rewiring process aims to improve GNN performance, particularly in tasks involving long-range dependencies. Several spatial graph rewiring techniques have been proposed to mitigate the over-squashing issue.

One of the earliest approaches to address over-squashing in GNNs was introduced by Alon and Yahav \cite{alo_yah-20a}. They identified that GNNs struggle with over-squashing in prediction tasks requiring extensive node interactions. As the number of layers increases, the receptive field of each node expands exponentially, making it difficult for information to propagate effectively. To counter this, they proposed a Fully-Adjacent (FA) layer. This method modifies only the final layer in an \(r\) layer GNN, replacing it with an FA layer that directly connects every pair of nodes with an edge. Converting an existing layer \eqref{eq:gnnlayer} to be fully-adjacent means that for every node \(u \in V\), \(N_u := V\) in that layer only. At layer \(r\), the transformation can be expressed as:

\begin{align}\label{eq:FAgnnlayer}
	h^{(r)}_u & = UP_{r}\left\{h^{(r-1)}_u, AGG_{r}\{h^{(r-1)}_v \text{ where } v \in V\}\right\}.
\end{align}
This transformation retains the original sparse structure in earlier layers while introducing dense connectivity only in the final layer. As a result, the FA layer improves long-range information propagation without altering earlier graph layers. Li et al. \cite{li-li-zha_24a} introduced a novel rewiring technique called Node-to-Node Distance Relationships (NNDR), designed specifically to mitigate over-squashing. The NNDR method computes higher powers of the adjacency matrix with self-loops, where each entry reflects the number of paths of a given length between node pairs. By analyzing these values, the method identifies nodes prone to over-squashing and strategically introduces new edges to improve information flow. To manage computational complexity, a sampling strategy ensures the rewiring process remains scalable for large graphs. To further enhance the model’s efficiency, the authors introduced a modified GNN architecture called Ordered Neurons for GNNs (O-GNN). Unlike standard relational GNNs (R-GNNs), which aggregate information from all edge types simultaneously, O-GNN distinguishes between original and newly added edges. This separation allows the model to prioritize information from nearby nodes while selectively preserving critical long-range connections. By combining NNDR-based rewiring with O-GNN’s structured learning approach, this method effectively addresses over-squashing while maintaining scalability.

Topping et al. \cite{top_gio_cha-21a} introduced a novel approach that utilizes graph curvature to address the over-squashing problem in GNNs. Their method, known as Stochastic Discrete Ricci Flow (SDRF), aims to surgically modify negatively-curved edges in order to reduce bottlenecks without significantly altering the statistical properties of the input graph. The core idea of SDRF is to target edges with high negative curvature, which are identified as key contributors to over-squashing. The SDRF algorithm operates in two main steps. First, it identifies edges with minimal Ricci-curvature\footnote{Given a graph \( G = (V, E) \) with vertices \( V \) and edges \( E \), the Ricci curvature of an edge \( e = (u, v) \) is denoted as \( \text{Ric}_e \) and is defined in terms of the ratio of the effective resistance between the endpoints of the edge and the length of the edge itself. It can be calculated using various methods, such as discrete Laplacian operators or spectral methods, for details see \cite{lin-lu-yan_11a}.}, indicating potential over-squashing issues. Then, it constructs additional supportive edges around these identified edges to reinforce their structural context within the graph. By strengthening these edges, the algorithm enhances the information flow within the GNN and mitigates the adverse effects of over-squashing. 
Black et al. \cite{bla_wan_nay-23a} conducted an in-depth analysis of over-squashing by exploring its connection with effective resistance between node pairs, as discussed previously in Section \ref{subsec:osqanalysis}. They proposed a novel graph rewiring method known as the Greedy Total Resistance (GTR) technique to address this issue. The GTR technique aims to minimize the effective resistance between nodes by strategically rewiring the graph. They derived a formula quantifying the impact of adding a specific edge \( (u,v) \) on the reduction of total resistance. The total change in resistance when adding an edge \( (u, v) \) to a connected graph \( G \) with \( n \) vertices can be expressed as the difference between the total resistance before and after the addition:
\[
R_{\text{tot}}(G) - R_{\text{tot}}(G \cup \{u, v\}) = n \cdot \frac{B_{u,v}^2}{1 + R_{u,v}}
\]
Here, \( n \) represents the number of vertices in the graph, \( B_{u,v} \) is the biharmonic distance of the node pair defined as:
\[
B_{u,v} = \sqrt{\left( \mathbf{1}_u -  \mathbf{1}_v \right)^T \left(\tilde{L}^+\right)^2 \left(  \mathbf{1}_u - \mathbf{1}_v \right)}
\]
and \( R_{u,v} \) denotes the effective resistance, as given in Equation \eqref{eq:eff_resistance}. The GTR method operates on a greedy principle, strategically adding edges to the graph to maximize \( \frac{B_{u,v}^2}{1 + R_{u,v}} \) and consequently minimize total resistance.

Topping et al. \cite{top_gio_cha-21a} identified over-squashing as a phenomenon associated with the presence of edges with negative curvature. In contrast, Black et al. and Di Giovanni \cite{bla_wan_nay-23a, gio_giu_far-23a} linked over-squashing to effective resistance or commute time between nodes, offering an alternative perspective on how message-passing bottlenecks arise in graph neural networks (GNNs).  

Taking a different approach, Gabrielsson et al. \cite{gab_yur_sol-22a} introduced a novel rewiring technique inspired by transformers, utilizing positional encoding to expand the receptive field of each node. Unlike curvature- or resistance-based methods, their model-agnostic approach enhances the original graph by adding nodes and edges while preserving its structure as a subset of the new rewired graph. Given a graph \(G\) and an integer \(r\), their method constructs a new graph \(G_r\) by connecting all nodes within \(r\)-hops and introducing a fully connected CLS (classification) node that links to every node in \(G_r\). However, a large \(r\) can lead to excessive densification, potentially distorting the original topology, which is crucial for graph-based learning. To counteract this, they embed positional encodings into \(G_r\), retaining information about the structure of \(G\) while enabling more effective message passing. These encodings, derived from random walk measures and the eigenvectors of the Laplacian matrix \cite{ram-gal_22a}, serve as node and/or edge features to maintain structural integrity.  

Another model-agnostic strategy for mitigating over-squashing is proposed by Attali et al. \cite{att-bus-per_25a}, who leverage edge curvature to guide information flow dynamically. Their curvature-guided rewiring framework restructures the graph based on edge curvature, enhancing message passing without modifying the underlying message-passing neural network (MPNN) architecture. By introducing a curvature-based homophily metric, the method selectively propagates information along edges with specific curvature properties, distinguishing between positively and negatively curved edges. This flexible approach improves connectivity while reducing bottlenecks. Furthermore, the framework integrates both one-hop and two-hop curvature-based propagation, ensuring efficient long-range communication without unnecessary graph densification. This structured rewiring method not only enhances information diffusion but also remains computationally efficient, making it a promising solution for improving GNN performance.


Another approach that probabilistically rewires the graph based on the given prediction task is the Probabilistically Rewired Message-Passing Graph Neural Network (PR-MPNN) \cite{qia-man_23a}. PR-MPNN leverages recent advancements in exact and differentiable k-subset sampling \cite{ahm-zen_22a} to probabilistically rewire the graph during the learning process. By doing so, PR-MPNN learns to add edges that are relevant to the prediction task while omitting less beneficial ones, effectively addressing issues like over-squashing and under-reaching.
Gutteridge et al. \cite{gut_don_bro-23a} propose a novel layer-dependent rewiring technique to address challenges posed by existing rewiring approaches \cite{abb-rad_22a, gab_yur_sol-22a} in GNNs. They highlight that some rewiring methods, while aiming to improve connectivity for long-range tasks, compromise the inductive bias provided by graph distance. These approaches enable instant communication between distant nodes at every layer, disrupting the inherent structure. To overcome this, Gutteridge and colleagues introduce a layer-dependent rewiring strategy, dynamically rewired message-passing with delay (DRew)  that gradually densifies the graph. This technique allows for enhanced connectivity while preserving the inductive bias provided by graph distance. Additionally, they incorporate a delay mechanism that facilitates skip connections based on both node distance and layer. This mechanism ensures that the graph's inductive bias is retained, providing a more nuanced and context-aware rewiring approach for improved performance in long-range tasks.

\begin{table*}[ht]
	\centering
	\caption{Taxonomy of Methods for Mitigating Limitations in GNNs Stemming from Long-Range Propagation. We categorize these approaches into two primary domains: spectral and spatial methods. Spectral methods in GNNs harness graph eigenvalues and eigenvectors for signal analysis and propagation, while spatial methods involve direct processing of node and edge information to capture local graph structures.
	 }\label{tab:models}
	\begin{tabular}{lllcccccc}
		\hline\\ [-0.7em]
		\multirow{2}{*}{Category} & \multirow{2}{*}{Approach} & \multirow{2}{*}{Methods} & \multicolumn{2}{c}{Task} & \multicolumn{2}{c}{Targets} & \multicolumn{2}{c}{Graph Type} \\
		\cline{4-9} \\[-0.7em]
		& & & Nodes & Graphs & Over-smoothing & Over-squashing & Homophily & Heterophily \\
		\hline \\[-0.7em] \vspace{.1cm}
\fontsize{7}{2}\selectfont		Spatial Method & Fully adjacent Rewiring  & FA \cite{alo_yah-20a} & \checkmark &\checkmark &  & \checkmark & & \\\vspace{.1cm}
\fontsize{7}{2}\selectfont		Spatial Method &\fontsize{7}{2}\selectfont Node-Node distance based rewiring & NNDR \cite{li-li-zha_24a} & & \checkmark& & \checkmark &\checkmark &\checkmark \\\vspace{.1cm}
\fontsize{7}{2}\selectfont		Spatial Method &\fontsize{7}{2}\selectfont Curvature-based rewiring & SDRF \cite{top_gio_cha-21a} & \checkmark & \checkmark& & \checkmark &\checkmark &\checkmark \\\vspace{.1cm}
\fontsize{7}{2}\selectfont		Spatial Method &\fontsize{7}{2}\selectfont Total Resistance & GTR \cite{bla_wan_nay-23a} & &\checkmark & &\checkmark&&\\ \vspace{.1 cm}
\fontsize{7}{2}\selectfont		Spatial Method & \fontsize{7}{2}\selectfont Probabilistic rewiring & PR-MPNN \cite{qia-man_23a} & &\checkmark&&\checkmark&&\\ 
\vspace{.1cm}
\fontsize{7}{2}\selectfont		Spatial Method &\fontsize{7}{2}\selectfont Layer-dependent Rewiring  & DRew \cite{gut_don_bro-23a} & \checkmark &\checkmark & & \checkmark & & \\\vspace{.1cm}
\fontsize{7}{2}\selectfont Spatial Method		&\fontsize{7}{2}\selectfont Training-free reservoir model & GESN \cite{tor_mic-22a} & \checkmark & & & \checkmark & &\checkmark \\ \vspace{.1 cm}
\fontsize{7}{2}\selectfont		Spatial Method& \fontsize{7}{2}\selectfont Curvature-based rewiring & BORF \cite{ngu_hie-23a} &\checkmark &\checkmark & \checkmark&\checkmark&\checkmark&\checkmark\\ \vspace{.1 cm}
\fontsize{7}{2}\selectfont		Spatial Method& \fontsize{7}{2}\selectfont Curvature-based rewiring & CCMP \cite{ngu_hie-23a} &\checkmark & & &\checkmark&\checkmark&\checkmark\\ \vspace{.1 cm}
\fontsize{7}{2}\selectfont		Spatial Method& \fontsize{7}{2}\selectfont Curvature-based rewiring & AFRC \cite{fes-web_24a} &\checkmark &\checkmark & \checkmark&\checkmark&\checkmark&\checkmark\\
\hline \\[-0.7em] \vspace{.1cm}
\fontsize{7}{2}\selectfont Spectral Method	& Expander graph & EGP \cite{dea_lac_vel-22a} & & \checkmark &&\checkmark &\\ \vspace{.1cm}
\fontsize{7}{2}\selectfont Spectral Method	&\fontsize{7}{2}\selectfont Commute time based rewiring & DiffWire \cite{arn_and_beg-22a} & \checkmark & \checkmark&  & \checkmark &\checkmark&\checkmark\\ \vspace{.1cm}
\fontsize{7}{2}\selectfont Spectral Method		&\fontsize{7}{2}\selectfont Curvature-based Edge flip & G-RLEF \cite{ban_kar_wan-22a} & &\checkmark & & \checkmark &&\\ \vspace{.1cm}
\fontsize{7}{2}\selectfont Spectral Method		& Edge addition  & FoSR \cite{kar_ban-22a} &  &\checkmark & \checkmark & \checkmark & & \\ \vspace{.1cm}
\fontsize{7}{2}\selectfont Spectral Method	&\fontsize{7}{2}\selectfont Edge addition \& removal   & SJLR \cite{gir_mal_bou-22a} & \checkmark & & \checkmark & \checkmark &\checkmark &\checkmark \\\vspace{.1cm} 
\fontsize{7}{2}\selectfont		Spectral Method &\fontsize{7}{2}\selectfont Curvature-based edge drop & CurvDrop \cite{liu_zho_pan-23a} & \checkmark &   \checkmark & \checkmark &\checkmark\\
\hline \\[-0.7em] \vspace{.1cm}
\fontsize{7}{2}\selectfont	Mixed Method &\fontsize{7}{2}\selectfont Locality-aware sequential rewiring & LASER \cite{fed-ame-ame_24a} & &   \checkmark & &\checkmark &\checkmark&\checkmark\\
\hline \\[-0.7em] \vspace{.1cm}
\fontsize{7}{2}\selectfont	Other Method&\fontsize{7}{2}\selectfont Directional Propagation&DGN\cite{bea_pas_vin-21a} & &\checkmark&\checkmark&\checkmark&& \\ \vspace{.1cm}
\fontsize{7}{2}\selectfont	Other Method&\fontsize{7}{2}\selectfont Heat kernel as filter & MHKG \cite{sha_shi_and-23a} &\checkmark&&\checkmark&\checkmark&\checkmark&\checkmark\\\vspace{.1cm} 
\fontsize{7}{2}\selectfont	Other Method&\fontsize{7}{2}\selectfont Graph ODEs &A-DGN \cite{gra_bac_gal-22a} & &\checkmark&&\checkmark&\checkmark&\checkmark \\ \vspace{.1cm}
\fontsize{7}{2}\selectfont		Other Method & \fontsize{7}{2}\selectfont Curvature-based pooling & CurvPool \cite{san_rot_lie-23a} & &\checkmark&\checkmark&\checkmark&&\\ \vspace{.1cm}
\fontsize{7}{2}\selectfont	Other Method &\fontsize{7}{2}\selectfont Topological information integration & PASTEL \cite{sun_li_yua-22a} &\checkmark&&&\checkmark&\checkmark&\checkmark\\ \vspace{.1cm}
\fontsize{7}{2}\selectfont	Other Method &\fontsize{7}{2}\selectfont Elementwise maximization & MGC \cite{she-qin-zha_24a} &\checkmark&&\checkmark&\checkmark&\checkmark&\checkmark\\ \vspace{.1cm}
\fontsize{7}{2}\selectfont		Other Method & \fontsize{7}{2}\selectfont Structural information propagation & RFGNN \cite{che_zha_li-22a} & &\checkmark&&\checkmark&&\\ 
		\hline
	\end{tabular}
\end{table*}

\subsection{Spectral Graph Rewiring Methods:}\label{subsec:spectralrewiring} To explain graph rewiring in the context of spectrum of the graph we would like to explain the connectedness of a graph with eigen values of the graph Laplacian. The connectedness of a graph $G$ can be measured via a quantity known as the Cheeger constant, denoted as $h(G)$, is defined as follows \cite{chu-97a}:
\begin{equation*}
	h(G) = \min_{(U\subset V)} \frac{\big|\{(u, v) \in E : u \in U, v \in V \setminus U\}\big|}{ \min\big(vol(U), vol(V \setminus U)\big)}
\end{equation*}
Here, $vol(U)$ represents the volume of set $U$ and is calculated as the sum of degrees of nodes $u\in U$.

The Cheeger constant, $h(G)$, essentially quantifies the energy required to divide graph $G$ into two separate communities. A smaller $h(G)$ implies that $G$ tends to have two communities with only a few connecting edges. In such cases, over-squashing is more likely to occur when information needs to traverse from one community to another. It's important to note that while computing $h(G)$ is generally a complex task, the Cheeger inequality provides a useful relationship: $h(G)$ is approximately proportional to the smallest positive eigenvalue \(\lambda_1\) of the graph Laplacian (spectral gap of the graph \cite{chu-97a}). We are assuming \(0=\lambda_0\le \lambda_1\le\cdots\lambda_n\) are the eigenvalues of the normalized graph Laplacian matrix \({L}\). 

In light of this relationship, some recent approaches have proposed selecting a rewiring strategy that depends on the spectrum of $G$. The goal is to generate a new graph $R(G)$ that satisfies $h(R(G)) > h(G)$. This strategy has been explored in the works \cite{arn_and_beg-22a, dea_lac_vel-22a, kar_ban-22a}. The underlying assumption is that propagating messages over the rewired graph $R(G)$ can mitigate over-squashing. 
However, it's important to note that this claim lacks formal analytical proof at this stage. 


Deac et al. \cite{dea_lac_vel-22a} proposed the Expander Graph Propagation (EGP) model for the graph classification task. In this model, information propagation occurs on expander graphs within an MPNN framework. Deac emphasized that for certain tasks, such as graph classification, relying solely on local node-level interactions is often insufficient for accurate predictions. In these cases, incorporating the global structure of the graph is equally important. Simply stacking additional message-passing layers on the original graph may fail to effectively capture such global information, especially in large graphs where bottlenecks hinder efficient communication. To address this, the EGP model employs expander graphs — a special class of sparse graphs with unique connectivity properties. Expander graphs are characterized by having a number of edges proportional to the number of nodes, i.e., \(|E| = O|V |\). This notation emphasizes that while expander graphs maintain sparsity, they achieve remarkably efficient connectivity. Despite having relatively few edges, expander graphs exhibit a low diameter property, meaning any two nodes can be connected in only a few hops. This ensures rapid and efficient information flow across distant nodes, effectively mitigating bottlenecks and reducing the risk of oversquashing. Consequently, the EGP model leverages these properties to improve message propagation in tasks that require both local and global structural awareness.
\begin{definition}\label{def:expandergraph}
A family \( \{ G_i \} \) of finite connected graphs is termed an expander family if there exists a constant \( c \geq 0 \) such that for every \( G_i \) in the family, the first non-zero eigenvalue \( \lambda_1(G_i) \) satisfies \( \lambda_1(G_i) \geq c \).
\end{definition}
Expander graphs can be interpreted in terms of Cheegers constant as: \cite{dea_lac_vel-22a} 
\begin{theorem}
Consider an infinite collection \( \{ G_i \} \) of expander graphs with a uniform upper bound on their vertex degrees. Then, there exists a constant \( \epsilon > 0 \) such that for all graphs in the collection, \( h(G_i) \geq \epsilon \).
\end{theorem}
Since expander graphs have higher Cheeger constants and will hence experience less severe problems
arising due to bottleneck edges
They proposed a method to efficiently construct expander of vertex approximately equal to \(|V|\) using special linear group.
\begin{definition}\label{def:slgroup}
For any positive integer \( n \), the special linear group \( \text{SL}(2, \mathbb{Z}_n) \) represents the group of \( 2 \times 2 \) matrices whose entries are integers modulo \( n \) and with determinant 1. One of its generating sets is:
\[ S_n = \left\{ \begin{pmatrix} 1 & 1 \\ 0 & 1 \end{pmatrix}, \begin{pmatrix} 1 & 0 \\ 1 & 1 \end{pmatrix} \right\}. \]
\end{definition} 
\begin{theorem}
The Cayley graph family \( \text{Cay}(SL(2, \mathbb{Z}_n); S_n) \) constitutes an expander family \cite{sel_65a} with vertices equal to the number of elements in \(SL(2, \mathbb{Z}_n)\).
\end{theorem}
Once the expander graph with appropriate \(n\) is constructed, EGP model follow a simple approach by running standard GNN layer on input structure \textit{i.e.} \(X,A\) followed by a GNN layer over the relevant Cayley graph.  Suppose we define \( A_{\text{Cay}}(n) \) as an adjacency matrix obtained from the Cayley graph \( Cay(SL(2, \mathbb{Z}_n); S_n) \). This leads to the expression:
\[ H = \text{GNN}(\text{GNN}(X, A), A_{\text{Cay}(n)}) \]
The EGP model is purposefully designed to enhance connectivity for long-range tasks, thereby facilitating efficient communication between distant nodes.

Arnaiz et al. \cite{arn_and_beg-22a} introduced a novel rewiring methodology that synthesizes the concepts of commute time and graph spectral gap. They exploit the Lovász bound:
\begin{equation}\label{eq:lovaszbound}
\left| \frac{\text{Com}(u,v)}{\text{vol}(G)}  - \left( \frac{1}{d_u} + \frac{1}{d_v} \right) \right| \leq \frac{1}{\lambda_1}\frac{2}{d_{\text{min}}},
\end{equation}
where \(\lambda_1 \geq 0\) represents the first non-zero eigenvalue of the normalized graph Laplacian matrix \(\tilde{L}\). This methodology comprises two distinctive layers within a Graph Neural Network (GNN) based in each side of the inequality \eqref{eq:lovaszbound}. The first layer, focuses
on the left-hand side, the Commute Time Layer (CT-LAYER), serves as a differentiable, parameter-free component tailored to learn the commute time. Meanwhile, the second layer, focuses on the right-hand side, the Gap Layer (GAP-LAYER), functions as the rewiring layer, tasked with optimizing the spectral gap based on the specific network characteristics and task objectives. The CT-LAYER calculates the commute times between nodes and retains edges with significant effective resistance \(R_{u,v}\) (\(=\frac{\text{Com}(u,v)}{\text{vol}(G)}\)) to preserve graph's topology. Subsequently, the GAP-LAYER adjusts the graph's adjacency matrix \(A\) utilizing ratio-cut and normalized-cut approximations \cite{buh-hei_09a} to minimize the spectral gap. During training, both CT-LAYER and GAP-LAYER dynamically learn the weights to predict optimal topology modifications for unseen graphs during testing. This integrated framework endows the GNN with the capability to adaptively learn and implement rewiring strategies, effectively mitigating challenges associated with over-squashing while considering the intricacies of graph structure and task requirements. Banerjee et al. \cite{ban_kar_wan-22a} introduced two graph rewiring techniques, namely Random Local Edge Flip (RLEF) and Greedy Random Local Edge Flip (G-RLEF), to alleviate bottlenecks in global information propagation within a graph. RLEF draws inspiration from the flip Markov chain \cite{all-bha-lat_16a, fed-gue-ada_06a, coo-dye-gre_19a}, which transforms a connected graph into an expander graph with high probability. For G-RLEF, the authors utilized the relationship between effective resistance \(R_{u,v}\) (Eq. \eqref{eq:eff_resistance}) and the number of triangles, denoted by \(\sharp_\Delta(u, v) = |N_u \cap N_v|\), that contain the edge \((u,v)\):
\begin{equation}\label{eq:Ruvno.oftriangles}
R_{u,v} \le \frac{2}{2+\sharp_\Delta(u, v)}
\end{equation}
They noted that a decrease in the number of triangles is accompanied by an increase in the spectral gap. Therefore, they devised a strategy to flip the edges \((u, i)\) and \((v, j)\) in a manner that minimizes the net change in the number of triangles. G-RLEF, a greedy variant of the RLEF algorithm, aims to expedite the spectral expansion process by employing a non-uniform sampling strategy for the hub edge \((u, v)\), where the selection is proportional to their effective resistance.

The key distinction between spatial and spectral rewiring lies in how they address bottlenecks in graph structures, particularly in mitigating oversquashing. Spatial rewiring techniques primarily connect nodes within a defined \(k\)-hop neighborhood \cite{gab_yur_sol-22a, abb-rad_22a}, or in extreme cases, operate over a fully-connected graph weighted by attention, as seen in Graph-Transformers \cite{kre_bea_ham-21a, ram-gal_22a}. While spatial rewiring partially preserves the graph’s locality by restricting connections to nearby nodes or incorporating positional information, these methods often result in denser computational graphs. This increased density raises memory complexity and may inadvertently exacerbate over-smoothing \cite{oon_suz-19a, gio_rus_bro-23a}. In contrast, spectral rewiring techniques improve graph connectivity by optimizing graph-theoretic measures related to expansion properties, such as the spectral gap, commute time, or effective resistance \cite{arn_and_beg-22a, kar_ban-22a, bla_wan_nay-23a}. These methods typically introduce fewer edges based on specific optimization criteria, preserving the graph’s sparsity. However, this approach disrupts the locality of the original graph by adding long-range edges between distant nodes. Consequently, while spectral rewiring can enhance connectivity, it may be less effective in addressing oversquashing due to its limited focus on local bottlenecks. To bridge the gap between these two approaches, Barbero et al. \cite{fed-ame-ame_24a} proposed the Locality-Aware SEquential Rewiring (LASER) framework. LASER combines the advantages of both spatial and spectral methods by selectively sampling edges to add based on equivariant, optimal conditions. This strategy maintains the inductive bias of spatial rewiring while ensuring the efficiency of spectral techniques, offering a more balanced and edge-efficient solution for mitigating oversquashing.

While spectral methods often rely on global metrics such as the spectral gap, these measures fail to provide insights into the precise locations of local bottlenecks. In contrast, discrete curvature measures offer a localized and computable metric to assess oversquashing directly from the graph’s topology. Building on this idea, Tori et al. \cite{tor-hol_24a} explored the effectiveness of curvature-based rewiring in improving performance on real-world graph datasets. Their findings align with Tortorella’s work \cite{tor_mic-23a}, which evaluated curvature-based rewiring in training-free GNNs and concluded that it seldom provides meaningful benefits for message passing. Tori et al. further demonstrated that curvature-based rewiring often fails to improve performance in real-world datasets. Their analysis attributed previously reported state-of-the-art (SOTA) results to hyperparameter tuning rather than the inherent benefits of the rewiring process itself — a pattern also noted in prior GNN performance studies \cite{ton-mar_24a}.

\section{Unifying Approaches to Address Over-Squashing and Over-Smoothing Trade-offs in Graph Neural Networks}\label{sec:osqosm} While these rewiring methods aim to enhance graph connectivity, they come with certain drawbacks, particularly when excessive modifications are made to the input graph. One prominent concern is the loss of valuable topological information inherent to the original graph. Additionally, the act of adding edges has a smoothing effect on the graph. If we introduce an excessive number of edges to the input graph, a standard Graph Convolutional Network (GCN) may encounter a common issue known as over-smoothing, as highlighted by Li et al. \cite{li_han__wu-18a}. 

\textit{Over-smoothing} in graph neural networks (GNNs) occurs when the embeddings of nodes from different classes become increasingly similar or indistinguishable \cite{hoa-mae-mur_21a, li_han__wu-18a}. This phenomenon is particularly prevalent in multi-layer MPNNs designed for short-range tasks, where a node's accurate prediction heavily relies on information from its immediate neighborhood. As the network's depth increases, the embeddings tend to lose the ability to differentiate between nodes of different classes due to the repeated aggregation of local information. In contrast,
\textit{over-squashing} in graph neural networks (GNNs) occurs when the number of layers in the network increases to accommodate long-range tasks, the receptive field of each node expands to include more neighboring nodes \cite{alo_yah-20a,top_gio_cha-21a}. Consequently, the information gathered from these extended receptive fields is compressed into fixed-length node vectors, resulting in a loss of detailed information from distant nodes. This compression of information leads over-squashing, where the network fails to accurately convey messages originating from distant nodes. 

Several methods address the intertwined challenges of over-smoothing and over-squashing in GNNs. Attali et al. \cite{att-bus_24a} provide a detailed review of state-of-the-art rewiring techniques, focusing on their theoretical foundations, practical implementations, and trade-offs, with an emphasis on improving GNN performance in complex settings like heterophilic graphs.



\begin{table*}[ht]
	\centering
	\caption{Critical Analysis of Spatial Graph Rewiring Methods}
	\label{tab:spatial_analysis}
	\begin{tabular}{|p{4cm}|p{6cm}|p{6cm}|}
		\hline
		\textbf{Method Category} & \textbf{Pros} & \textbf{Cons} \\ 
		\hline
		\textbf{Fully-Adjacent (FA) Layer} & 
		Improves long-range information propagation by introducing dense connectivity in the final layer. Retains the original sparse structure in earlier layers, preserving locality. & 
		Dense connectivity in the final layer increases computational complexity. May lead to over-smoothing due to excessive edge additions. \\
		\hline
		\textbf{Node-to-Node Distance Relationships (NNDR)} & 
		Strategically introduces new edges to improve information flow. Scalable due to sampling strategies for large graphs. & 
		Computationally expensive due to higher powers of adjacency matrix calculations. \\
		\hline
		\textbf{Stochastic Discrete Ricci Flow (SDRF)} & 
		Targets edges with high negative curvature, reducing bottlenecks. Enhances information flow without significantly altering the graph's statistical properties. & 
		Requires identification of negatively curved edges, which can be computationally intensive. \\
		\hline
		\textbf{Greedy Total Resistance (GTR)} & 
		Minimizes effective resistance between nodes, improving connectivity. Strategically adds edges to maximize reduction in total resistance. & 
		Greedy approach may not always find the optimal set of edges. \\
		\hline
		\textbf{Positional Encoding Rewiring} & 
		Expands the receptive field of each node using positional encodings. 
		Preserves the original graph structure while enabling more effective message passing. & 
		Large hop counts can lead to excessive densification, distorting the original topology. 
		Positional encodings may not always capture the necessary structural information. \\
		\hline
		\textbf{Curvature-Guided Rewiring} & 
		Dynamically restructures the graph based on edge curvature, enhancing message passing. 
		Reduces bottlenecks by selectively propagating information along specific curvature properties. & 
		Computationally expensive due to curvature calculations. 
		Requires careful hyperparameter tuning to balance between positive and negative curvature edges. \\
		\hline
		\textbf{Probabilistically Rewired Message-Passing (PR-MPNN)} & 
		Learns to add task-relevant edges while omitting less beneficial ones. 
		Addresses both over-squashing and under-reaching. & 
		Probabilistic rewiring may introduce instability during training. 
		Requires differentiable k-subset sampling, which can be computationally intensive. \\
		\hline
		\textbf{Layer-Dependent Rewiring (DRew)} & 
		Gradually densifies the graph, preserving the inductive bias of graph distance. 
		Skip connections based on node distance and layer improve long-range communication. & 
		Dynamic rewiring may increase memory complexity. 
		Requires careful design to balance between connectivity and preserving graph structure. \\
		\hline
	\end{tabular}
\end{table*}
\begin{table*}[ht]
	\centering
	\caption{Critical Analysis of Spectral Graph Rewiring Methods}
	\label{tab:spectral_analysis}
	\begin{tabular}{|p{4cm}|p{6cm}|p{6cm}|}
		\hline
		\textbf{Method Category} & \textbf{Pros} & \textbf{Cons} \\
		\hline
		\textbf{Expander Graph Propagation (EGP)} & 
		Uses expander graphs to enhance connectivity for long-range tasks. 
		Eliminates bottlenecks by ensuring short paths between nodes. & 
		Expander graph construction can be complex and computationally expensive. 
		May disrupt the locality of the original graph by adding long-range edges. \\
		\hline
		\textbf{Commute Time and Spectral Gap Rewiring} & 
		Optimizes the spectral gap to improve graph connectivity. 
		Dynamically learns weights to predict optimal topology modifications. & 
		Requires calculation of commute times and spectral gaps, which can be computationally intensive. 
		May not always preserve the original graph's local structure. \\
		\hline
		\textbf{Random Local Edge Flip (RLEF)} & 
		Transforms the graph into an expander graph with high probability. 
		Improves spectral expansion by flipping edges. & 
		Random flipping may not always lead to optimal connectivity. 
		May require multiple iterations to achieve desired spectral properties. \\
		\hline
		\textbf{Greedy Random Local Edge Flip (G-RLEF)} & 
		Expedites spectral expansion by non-uniform sampling of hub edges. 
		Reduces the number of triangles, increasing the spectral gap. & 
		Greedy approach may not always find the optimal set of edges to flip. 
		Computationally expensive due to the need to calculate effective resistance and triangle counts. \\
		\hline
		\textbf{Locality-Aware Sequential Rewiring (LASER)} & 
		Combines spatial and spectral rewiring advantages. 
		Selectively samples edges to add based on optimal conditions, preserving locality while enhancing connectivity. & 
		Requires careful design to balance between spatial and spectral rewiring. 
		May still introduce some computational overhead due to edge sampling and optimization. \\
		\hline
		\textbf{Curvature-Based Rewiring} & 
		Uses discrete curvature measures to assess over-squashing directly from graph topology. 
		Improves performance by targeting critical structures in the graph. & 
		Curvature-based rewiring often fails to improve performance in real-world datasets. 
		Computationally expensive and may require hyperparameter tuning. \\
		\hline
		\textbf{First-Order Spectral Rewiring (FoSR)} & 
		Optimizes the spectral gap by calculating the first-order change caused by adding edges. 
		Reduces over-smoothing by minimizing the Dirichlet energy. & 
		Requires calculation of the second eigenvector, which can be computationally intensive. 
		May not always preserve the original graph's local structure. \\
		\hline
		\textbf{Directional Graph Networks (DGN)} & 
		Propagates information in the direction of eigenvectors, reducing diffusion distance between nodes. 
		Effectively addresses over-squashing and over-smoothing. & 
		Requires calculation of eigenvectors, which can be computationally expensive. 
		May not be suitable for all graph structures. \\
		\hline
	\end{tabular}
\end{table*}
\begin{table*}[ht]
	\centering
	\caption{Critical Analysis of Other Rewiring Methods}
	\label{tab:other_analysis}
	\begin{tabular}{|p{4cm}|p{6cm}|p{6cm}|}
		\hline
		\textbf{Method Category} & \textbf{Pros} & \textbf{Cons} \\
		\hline
		\textbf{Maximization-based Graph Convolution (MGC)} & 
		Uses elementwise maximization to gather information from all powers of the adjacency matrix. 
		Improves information flow across distant nodes, addressing both over-smoothing and over-squashing. & 
		May sacrifice some long-range information to alleviate over-smoothing. 
		Requires careful design to balance between aggregation strategies and computational efficiency. \\
		\hline
		\textbf{DuoGNN} & 
		Decouples homophilic and heterophilic interactions, preventing over-smoothing.
		Captures long-range dependencies while maintaining a compact structure. & 
		Requires separate processing of homophilic and heterophilic graphs, increasing complexity.
		May require careful tuning to balance between different types of interactions. \\
		\hline
		\textbf{DeltaGNN} & 
		Dynamically removes edges with low Interaction Filtering and Clustering (IFC) scores. 
		Restores long-range dependencies by identifying key nodes with high IFC scores. & 
		Requires calculation of IFC scores, which can be computationally intensive.
		May require careful tuning to balance between edge removal and long-range dependency restoration. \\
		\hline
		\textbf{Graph Transformers} & 
		  Less susceptible to over-smoothing compared to traditional GNNs.
		  Establishes direct paths between distant nodes, alleviating over-squashing. & 
		  High computational and memory requirements due to global attention mechanisms.
		  May result in improper training due to blending of local and non-local interactions. \\
		\hline
		\textbf{Graph ViT/Mixer MLP} & 
		  Captures long-range dependencies while mitigating over-squashing.
		  Offers improved computational efficiency, speed, and memory advantages. & 
		  May not be suitable for all graph structures.
		  Requires careful design to balance between local and non-local interactions. \\
		\hline
		\textbf{Redundancy-Free GNN (RFGNN)} & 
		  Eliminates redundancy in message propagation using truncated ePaths trees.
		  Balances epath influence, enhancing GNNs' ability to capture structural information. & 
		  Requires construction of path-search-trees, which can be computationally intensive.
		  May require careful tuning to balance between epath influence and computational efficiency. \\
		\hline
		\textbf{Graph Echo State Network (GESN)} & 
		  Training-free, making it efficient for node classification tasks.
		  Mitigates issues of long-range message passing and over-squashing in heterophilic graphs. & 
		  Limited to node classification tasks. 
		  May not be suitable for all graph structures. \\
		\hline
		\textbf{Anti-Symmetric DGN (A-DGN)} & 
		  Preserves long-term dependencies between nodes using anti-symmetric weight matrices. 
		  Alleviates over-squashing by enforcing specific properties on the ODE system. & 
		  Requires careful design to ensure stability and non-dissipative properties.
		  May not be suitable for all graph structures. \\
		\hline
		\textbf{Multi-Scaled Heat Kernel GNN (MHKG)} & 
		  Combines smoothing and sharpening effects on node features, managing the trade-off between over-smoothing and over-squashing.
		  Uses curvature-based pooling to guide the pooling process. & 
		  Requires calculation of heat kernels and curvature values, which can be computationally intensive.
		  May require careful tuning to balance between smoothing and sharpening effects. \\
		\hline
		\textbf{Curvature-Based Pooling (CurvPool)} & 
		  Balances over-smoothing and over-squashing by pooling nodes with similar curvature profiles.
		  Retains the original graph structure by remapping old edges to new node clusters. & 
		  Requires calculation of curvature values, which can be computationally intensive.
		  May require careful tuning to balance between pooling and preserving graph structure. \\
		\hline
	\end{tabular}
\end{table*}

Nguyen et al in \cite{ngu_hie-23a} reveals a connection between local graph geometry and the issues of over-smoothing and over-squashing and introduced a novel rewiring technique known as Batch Ollivier-Ricci Flow (BORF), which harnesses the power of Ollivier-Ricci curvature to address these interrelated challenges in GNNs. Mathematically over-smoothing can be seen as 
\begin{equation}\label{eq:osm_borf}
	\sum_{(u,v)\in E}\left|h_u^{(k)}-h_v^{(k)}\right| \rightarrow 0 \text{ as } k \rightarrow \infty
\end{equation}
Over-smoothing increases as number of layer \(k\) increases as node features become indistinguishable. This definition is similar to the definition based on the node-wise Dirichlet energy 
\begin{equation}\label{eq:dirichletenergy}
\mathcal{E}(X) = \frac{1}{2} \sum_{i,j,k} A_{uv} \left( \frac{X_{uk}}{\sqrt{d_u}} - \frac{X_{vk}}{\sqrt{d_v}} \right)^2
\end{equation}
given in \cite{rus-cha_22a,kar_ban-22a} where $X$ is the feature matrix and $A$ is the adjacency matrix. In addition to the connection of edges with negative curvature as discussed in \cite{top_gio_cha-21a} also, Nguyen et al linked over-smoothing to edges with positive curvature. Specifically, they proved that, if every edge curvature in a regular graph \(G\) is bounded from below by a sufficiently high constant then the difference between the features of any pair of neighboring nodes , exponentially converges to \(0\) in a typical GNN, \textit{i.e}, 
\begin{equation}\label{eq:positivecur_osm}
	\sum_{(u,v)\in E}\left|h_u^{(k)}-h_v^{(k)}\right| \le c_1 e^{-c_2k},
\end{equation}
where \(c_1, c_2\) are positive constant.
BORF operates in batches and begins by identifying two sets of edges in each batch: $p$ edges with minimal curvature and $q$ edges with maximal curvature. 
It then optimizes the graph's connectivity by adding connections to the minimally curved edges, ensuring efficient communication between distant nodes. This alleviates over-squashing. 
BORF mitigates over-smoothing by removing edges with the highest curvature, as these can excessively smooth node features. The algorithm's adaptability allows it to function as a net edge addition, subtraction, or net-zero rewiring, making it flexible across different datasets.

In a similar vein, Liu et al. \cite{liu_zho_pan-23a} established a connection between over-smoothing and positively curved edges, as well as between over-squashing and negatively curved edges. They tackled both issues by employing curvature-based edge removal, introducing a sampling layer driven by Ricci curvature. This layer selectively removes edges with low Ricci curvature at each GNN layer, effectively mitigating both over-smoothing and over-squashing. However, Fesser and Weber \cite{fes-web_24a} highlight that many curvature-based methods rely on computationally expensive subroutines and require careful hyperparameter tuning, making them impractical for large-scale graphs. To address this, they propose an Augmented Forman-Ricci Curvature (AFRC)-based rewiring technique, which can be computed in linear time. Their method also includes an efficiently computable heuristic for automating hyperparameter selection, eliminating the need for costly tuning. AFRC effectively captures both over-smoothing and over-squashing effects in message-passing GNNs. Dai Shi et al. \cite{shi_guo_zhi-23a} introduced Curvature-Based Edge Dropping (CBED), a novel algorithm that strategically removes edges with the highest positive curvature. This approach improves the model’s adaptability to heterophilic graphs while simultaneously reducing over-smoothing, further advancing the effectiveness of curvature-based rewiring techniques in GNNs.


Giraldo et al. \cite{gir_mal_bou-22a} established a profound trade-off between over-smoothing and over-squashing. In the context where \( P = D^{-1}A \) is the random walk transition matrix, for any initial distribution \( f : V \rightarrow \mathbb{R} \) with \( \sum_{v \in V} f(v) = 1 \), the distribution after \( k \) steps is given by \( f^T P^k \), where \( f \in \mathbb{R}^{N \times 1} \) is the vector of initial distributions such that \( f(i) \) represents the function evaluated on the \( i \)-th node. The random walk is ergodic when there is a unique stationary distribution \( \pi \) satisfying \( \lim_{s \rightarrow \infty} f^T P^s = \pi \). They proved the following results
\begin{theorem}
Let \( h(G) \) be the Cheeger constant of \( G \), and let \( s \) be the number of required steps such that the \( \ell_2 \) distance between \( f^T P^s \) and \( \pi \) is at most \( \varepsilon \). Then
\[ 2h(G) \geq \frac{1}{s} \log \frac{{\max_{u\in V} \sqrt{d_u}}}{{\varepsilon \min_{v\in V} \sqrt{d_v}}} \]
\end{theorem}
That is, if \( s \) approaches 0, then \( h_G \) approaches infinity, implying that we can reduce the bottleneck effect in the graph by speeding up the convergence to the stationary distribution. Conversely, if \( h_G \) approaches 0, then \( s \) approaches infinity, indicating that we can avoid converging to the stationary distribution by promoting a bottleneck-like structure in the graph.
In response to this challenge, Giraldo et al. introduced the Stochastic Jost and Liu Curvature Rewiring (SJLR) algorithm, a notable departure from previous curvature-based techniques \cite{top_gio_cha-21a, ban_kar_wan-22a, kar_ban-22a}. 
SJLR dynamically adds and removes edges during the training phase of GNNs while maintaining the fundamental graph structure unaltered during the testing phase. This adaptability sets SJLR apart as a promising approach to address the intricate challenges posed by over-smoothing and over-squashing in GNNs. To tackle the trade-off between these challenges, Karhadkar et al. \cite{kar_ban-22a} proposed a novel rewiring method called  First-order Spectral Rewiring (FoSR) with the objective of optimizing the spectral gap of the graph input to the GNN. 

\begin{theorem}
	The first-order change in the second eigenvalue \( \lambda_1 = \lambda_1(D^{-1/2}AD^{-1/2}) \) resulting from adding the edge \( (u, v) \) is approximately equal to \cite{kar_ban-22a}:
	\begin{equation}\label{eq:IstorderchangeinSG}
		\frac{2\mu(u) \mu(v)}{(\sqrt{1 + d_u})(\sqrt{1 + d_v})},
	\end{equation} 
	where \( \mu \) represents the second eigenvector of \( D^{-1/2}AD^{-1/2} \), and \( \mu(u) \) denotes the \( u \)-th entry of \( \mu \).
\end{theorem}
The FoSR algorithm carefully calculates the first-order change in the spectral gap caused by adding each edge and then chooses the edge that maximizes this change. In other words, it approximates \(\mu\) and selects an edge that minimizes the expression given in Equation \eqref{eq:IstorderchangeinSG}. Within this framework, the authors propose a comprehensive approach that not only introduces this innovative rewiring method but also integrates a relational Graph Neural Network to effectively leverage these rewired edges. They have demonstrated that the Dirichlet energy \eqref{eq:dirichletenergy}, consequently the over-smoothing in the proposed GNN is lower than that of traditional GNNs. Beaini et al. in \cite{bea_pas_vin-21a} have associated these challenges with the incapacity of Graph Neural Networks (GNNs) to effectively capture directional information within graphs. This limitation constrains their ability to understand graph structures and perform feature transformations. To address this, they introduced Directional Graph Networks (DGNs), leveraging the eigenvectors \( \lambda_j \) of the normalized Laplacian matrix \( \tilde{L} \). They demonstrated that by propagating information in the direction of \( \mu \), DGNs efficiently facilitate information sharing between distant nodes in the graph, thereby reducing the diffusion distance between them.
\begin{theorem}
	Let \( u \) and \( v \) be two nodes in graph \( G \) such that \( \mu(u) < \mu(v) \). If \( u_0 \) is the node obtained by moving one step from \( u \) in the direction of \( \nabla \mu \), then there exists a constant \( C \) such that for \( C \leq t \), the diffusion distance \[ d_t(u_0, v) < d_t(u, v) .\] This reduction in distance is proportional to \( e^{-\lambda_1} \).
\end{theorem}
The diffusion distance at time \( t \) between nodes \( u \) and \( v \) is given by:
\[d_t(u,v) = \left(\sum_{u'\in V}\left(p_t(u,u')-p_t(u',v)\right)^2\right)^{\frac{1}{2}},\]
where \( p_t(u,v) = P(P^t = v | u_0 = u)\). In summary, the DGN model, through its globally consistent directional information, effectively addresses challenges such as over-squashing and over-smoothing. It empowers GNNs to comprehend local graph structures, perform meaningful feature transformations, and mitigate the adverse effects of these issues.

Several methods have been proposed to address over-smoothing and over-squashing in GNNs by aggregating features from multiple neighborhood hops using weighted combinations of adjacency matrix powers. Models such as SGC \cite{wu-sou-zha_19a}, S$^2$GC \cite{zho-hua-li_20a}, and APPNP \cite{gas_etal-19a} follow this approach, while GDC \cite{gas_etal-19b} generalizes the graph diffusion process to provide a unified framework. These methods extend GCN depth without severe performance degradation; however, they inevitably sacrifice some long-range information to alleviate over-smoothing, resulting in a tradeoff between mitigating over-smoothing and addressing over-squashing \cite{alo_yah-20a}.

To overcome this limitation, Shen et al. \cite{she-qin-zha_24a} proposed the Maximization-based Graph Convolution (MGC) model. Unlike traditional models that rely on linear aggregation of multi-hop neighborhood features, MGC employs an elementwise maximization operation to gather information from all powers of the adjacency matrix. This approach improves information flow across distant nodes, effectively addressing both over-smoothing and over-squashing. To improve scalability for large graphs, the authors introduced an approximated variant (AMGC) with linear complexity in the number of edges. MGC follows a decoupled design where feature propagation occurs in a preprocessing stage, followed by a simple MLP-based transformation during training. By combining original and processed node features, this architecture enhances model expressiveness while maintaining computational efficiency.

While MGC focuses on improving aggregation strategies, Mancini and Rekik \cite{man-isl_25a} address over-smoothing and over-squashing through an edge-filtering approach. Their proposed model, DuoGNN, introduces a scalable and generalizable GNN architecture that decouples homophilic and heterophilic interactions in a topological interaction-decoupling stage. This separation ensures that each type of interaction is processed independently. DuoGNN follows three key steps: First, a topological edge-filtering step removes heterophilic edges to construct a strongly homophilic subgraph \(G_{ho}\), preventing over-smoothing and preserving short-range interactions. Second, a heterophilic graph condensation step extracts important heterophilic interactions from homophilic clusters to form a new subgraph \(G_{he}\) which captures long-range dependencies while maintaining a compact structure. Finally, DuoGNN employs two independent GNN modules that process \(G_{ho}\) and \(G_{he}\) separately—one learning homophilic aggregation and the other modeling heterophilic interactions. This dual-path design enables DuoGNN to balance expressiveness and scalability effectively.

Building upon concepts from DuoGNN, DeltaGNN \cite{man-rek_25a} extends this idea by introducing a more structured Interaction Filtering and Clustering (IFC) mechanism. Similar to DuoGNN, DeltaGNN begins by performing homophilic aggregation to capture short-range dependencies. However, rather than relying on fixed connectivity rules, DeltaGNN dynamically removes edges with the lowest IFC scores, calculated using Euclidean distance. This targeted edge removal refines the graph structure by eliminating bottlenecks while preserving homophilic components. To restore long-range dependencies lost during this filtering, DeltaGNN employs a heterophilic graph condensation process similar to DuoGNN’s strategy. However, DeltaGNN identifies key nodes with high IFC scores to build this graph, ensuring efficient retention of critical long-range interactions. Both DuoGNN and DeltaGNN adopt a dual aggregation strategy, processing homophilic and heterophilic graphs independently. While DuoGNN emphasizes graph topology to split interactions, DeltaGNN’s adaptive filtering process enhances scalability and performance on complex datasets.

\section{Graph Transformers and other Strategies}\label{sec:otherarchitecture} Graph transformers have gained substantial attention as an alternative approach to combating over-smoothing and over-squashing in the context of graph and computer vision domains \cite{yun_jeo_kim-19a, che_wu_dai-22a, cai_lam-20a}. This approach leverages the inherent strengths of transformer architectures. for instance, Ying et al. \cite{yin_cai_luo-21a} observed that transformers are less susceptible to over-smoothing compared to traditional GNNs. Their ability to model graph data efficiently contributes to mitigating the over-smoothing problem. Kreuzer et al. \cite{kre_bea_ham-21a} highlighted the resilience of transformers to over-squashing. Transformers establish direct paths connecting distant nodes, which alleviates the over-squashing challenge.


However, it's worth noting that transformers have limitations, including significant computational and memory requirements due to the need for every node to attend to all others. This can make them less suitable for large-scale graph applications and may result in improper training leading to a blend of local and non-local interactions. To tackle this challenge several methods are proposed. 
\begin{table*}[ht]
	\centering
	\caption{Dataset Statistics}\label{tab:dataset}
	\begin{tabular}{lllcccccccccc}
		\hline\\ [-0.7em]
		\multirow{2}{*}{} & \multirow{2}{*}{} & \multirow{2}{*}{} &\multirow{2}{*}{} & \multirow{2}{*}{} &\multicolumn{2}{c}{\textbf{Node classification}} & \multirow{2}{*}{} & \multirow{2}{*}{} &&& \\
		\cline{1-13}\\[-0.7em]
		&Cora & Citeseer & Pubmed & Actor & TwitchDE & Tolokers &Cornell &Texas & Wisconsin & Chemeleon & Squirrel & Roman-Empire\\
		\hline \\[-0.7em] \vspace{.1cm}
		\#nodes & 2708 & 3327 & 19717 & 7600 & 9496 &11758 &183&183&251&2277&5201& 22662\\ \vspace{.1cm}
		\#edges& 10556 & 9104 & 88648& 26752 &153138&519000&280&295&466&31421&198493&32927\\ \vspace{.1cm}
		\#features & 1433  & 3703 & 500&932  & 2514 &10&1703&1703&1703&2089&2089&300\\ \vspace{.1cm}
		\#classes& 7  & 6&3 & 5 & 2 &2&5&5&5&5&5&18\\ \vspace{.1cm}
		$H(G)$& 0.81 & 0.74 &0.80 & 0.22 & 0.63 &0.59 & 0.30&0.11&0.21&0.23&0.22&0.06 \\ \vspace{.1cm}
		Directed  & $\times$ & $\times$ & $\times$& \checkmark & \checkmark & \checkmark &\checkmark&\checkmark&\checkmark&\checkmark&\checkmark&\checkmark\\ 
		\hline\\ [-0.7em]
		\multirow{2}{*}{} & \multirow{2}{*}{} &\multirow{2}{*}{} &\multirow{2}{*}{} &\multirow{2}{*}{} & \multicolumn{2}{c}{\textbf{Graph classification}} & \multirow{2}{*}{} & \multirow{2}{*}{} \\
		\cline{1-13} \\[-0.7em]
		&NCI-1 & NCI-109 & Reddit-B &Reddit-5K & Reddit-12K & Collab & Enzymes &BZR & MUTAG & PTC & COX2 & Proteins \\
		\hline \\[-0.7em] \vspace{.1cm}
		\#graphs & 4110 & 4127 & 200 & 4999 & 11929 &5000&600&405&188&344&467&1113\\ \vspace{.1cm}
		Avg. nodes & 30 & 30 &430 & 509 &391&75&32.63&35.75&17.93&25.56&41.22&39.06\\ \vspace{.1cm}
		Avg. edges &32 & 32 & 498& 595 & 457 &2458&62.14&38.36&19.79&25.96&43.44&72.82\\ \vspace{.1cm}
		\#features & 37  & 38 & 0& 0 & 0 &0&0&0&0&0&0&1\\ \vspace{.1cm}
		\#classes&2 & 2 &2 & 5 & 11&3&6&2&2&2&2&2\\ \vspace{.1cm}
		Directed & $\times$ &$\times$  &$\times$ &$\times$ &$\times$  &$\times$&$\times$&$\times$&$\times$&$\times$&\checkmark&$\times$  \\
		\hline
	\end{tabular}
\end{table*}

Xiaoxin \cite{he_hoo_lau-23a} introduces a novel approach as an alternative to global attention mechanisms. This approach draws inspiration from ViT\textbackslash \text{Mixer MLP} architectures initially introduced in computer vision. The resulting "graph ViT\textbackslash \text{Mixer MLP}" GNNs excel in capturing long-range dependencies while effectively mitigating over-squashing issues. They offer improved computational efficiency, speed, and memory advantages compared to existing models. Qingyun et al. in \cite{sun_li_yua-22a} address the challenge of over-squashing in Graph Neural Networks (GNNs) by emphasizing its correlation with topology imbalance. Introducing Position-Aware STructurE Learning (PASTEL) as a solution, they redefine topology imbalance in the context of under-reaching and over-squashing, establishing two quantitative metrics, reaching and squashing coefficients, respectively for assessment.
\begin{definition}
For a given graph $G$, the Reaching Coefficient (RC) measures the average length of the shortest path from unlabeled nodes to their corresponding labeled nodes within their respective classes:
\[RC =\frac{1}{|V^U|}\sum_{u\in V^U}\frac{1}{|V^L|}\sum_{v\in V^L} \left(1-\frac{\log d_G}{Diam(G)}\right),\]
where \(V_L\) and \(V^U\) denotes labeled and unlabeled nodes.
\end{definition} 
The reaching coefficient indicates the extent of information propagation from labeled to unlabeled nodes by GNNs. It's important to note that $RC \in [0, 1)$, where a higher \(RC\) signifies improved reachability
\begin{definition}
For a graph $G$, the Squashing Coefficient (SC) quantifies the average Ricci curvature of edges along the shortest path from unlabeled nodes to the labeled nodes within their corresponding classes:
\[SC = \frac{1}{|V^U|}\sum_{u\in V^U}\frac{1}{|N_u^L|}\sum_{v\in N_u^L}\frac{\sum_{(u',v')\in P_{uv}}Ric(u',v')}{d_G},\]
where \(N_u^L\) is the labeled neighborhood of node \(u\), and \(Ric(u,v)\) denotes the Ollivier-Ricci curvature \cite{oll_09a}.
\end{definition}
The squashing coefficient may exhibit either positive or negative values, with a larger \(SC\) indicating reduced squashing. PASTEL is designed to improve intra-class connectivity among nodes in GNNs by optimizing information propagation paths. This is achieved through the utilization of transformer based position encoding mechanism that captures the relative positions of unlabeled nodes with respect to labeled nodes. Furthermore, a class-wise conflict measure, employing Group PageRank, assesses the influence of labeled nodes from different classes, guiding adjustments to edge weights to enhance intra-class connectivity.
Rongqin et al. \cite{che_zha_li-22a} delve into the challenge of over-squashing within GNNs, linking it to message redundancy during aggregation. They establish that redundancy significantly contributes to over-squashing, especially when conventional GNNs struggle with long-length path propagation, limiting their ability to handle long-range interactions. To address this, they introduce the Redundancy-Free Graph Neural Network (RFGNN), leveraging extended paths (epaths)\footnote{A path that has no repeated node except for the first node as the end node in a path of length more than 2.}, to capture complex graph structures. They implement truncated ePaths trees (TPTs) for message-passing, where TPTs of height \(k\) (\(TPT_{(G,u)^k}\)) are obtained by running a BFS from node \(u\) in graph \(G\), accessing epaths of length up to \(k\). RFGNN employs a path-search-tree concept constructed via breadth-first search to eliminate redundancy in message propagation, ensuring efficient information transmission without over-squashing. This novel de-redundancy technique balances epath influence, enhancing GNNs' ability to capture structural information in original graphs while mitigating over-squashing. Tortorella and Mechelli \cite{tor_mic-22a} address the issue of over-squashing in node classification tasks within graphs characterized by low homophily (as defined in Equation \eqref{eq:graphhomophily}). They propose a reservoir computing model called the Graph Echo State Network (GESN).
\begin{definition}
	Reservoir computing \cite{nak_18a} presents a paradigm for designing recurrent neural networks (RNNs) with efficiency. Input data is encoded by a randomly initialized reservoir, with only the readout layer requiring training for downstream task predictions.
\end{definition}
GESN extends the reservoir computing paradigm to graph-structured data. Node embeddings are computed recursively through a nonlinear dynamical system represented by the equation:
\begin{equation}\label{eq:gesnebedding}
	h^{(k)}_u = \text{tanh} \left( W_{\text{in}} x_u + \sum_{v \in N(u)} W_{\hat{h}} h^{(k-1)}_{v} \right), \quad h^{(0)}_v = 0.
\end{equation}
Here, \( W_{\text{in}} \in \mathbb{R}^{n \times d_0} \) and \( W_{\hat{h}} \in \mathbb{R}^{n \times n} \) are the input-to-reservoir and recurrent weights, respectively, for a reservoir with \( n \) units. Equation \eqref{eq:gesnebedding} iterates over \( k \) until the system state converges to the fixed point \( h^{(\infty)}_u \), which serves as the embedding. For node classification tasks, a linear readout is applied to node embeddings. The training-free characteristic makes GESN an efficient and effective solution for node classification tasks, offering a promising approach to mitigate issues of long-range message passing and over-squashing in heterophilic graphs.

Gravina et al. in \cite{gra_bac_gal-22a} introduced the Anti-Symmetric Deep Graph Network (A-DGN), an innovative framework tailored to address the challenge of long-term information propagation in Deep Graph Networks (DGNs). This approach leverages principles from ordinary differential equations (ODEs) by representing node features as a function of time, \(x_u(t)\) for \(u\in V\), and defining a node-wise ODE as:
\begin{equation}\label{eq:nodeasode}
	\frac{dx_{u}(t)}{dt} = f_G(x_u(t)),
\end{equation}
for \(t\in [0,T]\), with initial conditions \(x_u(0) = x_u\in \mathbb{R}^{d_0}\). Here, \(f_G:\mathbb{R}^{d_0} \rightarrow \mathbb{R}^{d_0}\) represents the dynamics of node representations. This ODE process can be seen as continuous information processing over graphs, starting from initial features \(x_u(0)\) and culminating in final node features \(x_u(T)\). This process shares similarities with standard DGNs, as it computes nodes' states that can be used as an embedded representation of the graph and then fed into a readout layer for downstream tasks on graphs. Gravina et al. establish theoretical conditions under which a stable and non-dissipative ODE system can be realized on graph structures, utilizing anti-symmetric\footnote{A square matrix \(A\) is anti-symmetric if \(A^T = -A\).} weight matrices. The A-DGN layer is formulated through the forward Euler discretization of the obtained graph ODE:
\begin{multline}
x_u^\ell = x_u^{\ell-1}+\epsilon UP\left((W-W^T-\gamma I)x_u^{\ell-1}\right.\\
\left.+AGG(x_v(t), \text{ where } v\in N_u)\right),
\end{multline}
where \(\gamma\) controls the strength of diffusion and \(\epsilon\) is the discretization step. This process enforces specific properties on the ODE system, preserving long-term dependencies between nodes within the graph and alleviating the problem of over-squashing in GNNs. In their paper \cite{sha_shi_and-23a}, Shao et al. introduced a novel algorithm, the Multi-Scaled Heat Kernel based Graph Neural Network (MHKG), aiming to address the challenge of over-smoothing as well as over-squashing in GNNs. They propose a generalized graph heat equation represented as:
\begin{equation}\label{eq:generalizedgraphode}
	\frac{dH^{(t)}}{dt} = -f(\tilde{L})H^{(t)},
\end{equation}
with the initial condition $H^{(0)}=X \in \mathbb{R}^{n\times d_0}$, where $f(\tilde{L})$ acts as a filtering function operating element-wise on the eigenvalues of $\tilde{L}$, typically represented by polynomial or analytic functions. Mathematically, this is denoted as $f(\tilde{L}) = U f(\Lambda) U^T$, where $U$ is the matrix of eigenvectors and $\Lambda$ is the diagonal matrix of eigenvalues of $\tilde{L}$. Similar to the basic heat equation, they define $K_t = e^{-tf(\tilde{L})}$ as the generalized heat kernel. The ODEs \eqref{eq:generalizedgraphode} used in GNNs mimic heat flow in a fixed direction and speed, akin to the second law of thermodynamics, leading to over-smoothing, where all node features become equal over time. The proposed MHKG model introduces a mixed dynamic approach, combining both smoothing and sharpening effects on node features, formulated as:
\begin{multline}
	H^{(\ell)} = U \text{diag}(\theta_1) \Lambda_1 U^\top H^{(\ell-1)} W^{(\ell-1)}\\ + U \text{diag}(\theta_2) \Lambda_2 U^\top H^{(\ell-1)} W^{(\ell-1)},
\end{multline}
where $\text{diag}(\theta_1)$ and $\text{diag}(\theta_2)$ are trainable filtering matrices, and $\Lambda_1 = -f(\Lambda) = \text{diag}{e^{-f(\lambda_i)}}{i=1}^N$ and $\Lambda_2 = f(\Lambda) = \text{diag}{e^{f(\lambda_i)}}{i=1}^N$. The model manipulates the graph spectral domain through controlled adjustments of time, effectively enhancing node feature sharpness while managing the trade-off. Curvature based pooling technique is also proposed to deal with over-smoothing and over-squashing in GNNs. Curvature-based Pooling within Graph Neural Networks (CurvePool) \cite{san_rot_lie-23a} relies on the Balanced Forman curvature \cite{top_gio_cha-21a} to identify critical structures in the graph that contribute to these problems. This method calculates curvature values for each edge and employs a criterion to group nodes into clusters, ensuring that nodes with similar curvature profiles are pooled together. The resulting node clusters are transformed into new nodes in the pooled graph, and node representations within each cluster are aggregated using operators like mean, sum, or maximum. To retain the original graph structure, CurvPool remaps old edges to the new node clusters. By leveraging graph curvature to guide the pooling process, CurvPool effectively balances over-smoothing and over-squashing, ultimately improving the performance of GNNs in graph classification tasks. 

Each of these approaches offers a unique perspective and set of techniques to address the challenges of over-squashing in graph-based machine learning models.

\section{Datasets}\label{sec:datasets}
The common datasets employed for node and graph classification tasks in the models listed in Table \ref{tab:models} are presented in Table \ref{tab:dataset}, along with detailed dataset statistics. It's important to note that this list is not exhaustive, as there are numerous other datasets, including synthetic and large-scale real-world ones, utilized for various research purposes. Table \ref{tab:dataset} displays the statistics of the datasets used in this study, where $H(G)$ represents the graph's homophily, as defined in \cite{pei_wei_cha-20a}, calculated as
\begin{equation}\label{eq:graphhomophily}
H(G)= \frac{1}{|V|}\sum_{v\in V}\dfrac{\text{$v$'s neighbors with the same label as $v$}}{N_v}
\end{equation}

For node classification tasks, we employ a diverse set of 12 datasets, encompassing graphs of varying sizes and characteristics.

Cora \cite{mcc_nig_ren-00a}, CiteSeer \cite{sen_nam-08a}, and PubMed \cite{nam_lon-12a} are examples of paper citation networks. In these datasets, node features are represented as bag-of-words extracted from paper content, and the goal is to classify research topics. Notably, these datasets exhibit high homophily. In contrast, the Actor \cite{tan_sun_wan-09a} dataset is created based on actor co-occurrences on Wikipedia pages and is categorized into five groups. This dataset poses a node classification task with low homophily characteristics. 
TwitchDE, on the other hand, is a social network comprising German gamer accounts from Twitch, categorized as suitable for work or adult profiles. The classification task involves profiling these accounts. The Tolokers dataset represents a collaboration network derived from the crowdsourcing platform Toloka. The objective here is to determine user activity, considering the challenge of class imbalance, with the evaluation metric being the area under the ROC curve. Cornell, Texas, Wisconsin are additional node classification tasks, each originating from university-related interactions. The Cornell dataset comprises research paper citation data. The Texas dataset represents friendships in a Texas college, and the Wisconsin dataset is derived from a university-related network. Node features and specific targets for these datasets can vary. Chameleon\cite{roz_all_sar-21a}, Squirrel\cite{roz_all_sar-21a} and Roman-Empire \cite{pla-kuz_23a} are also novel datasets introduced for node classification. Chameleon captures interactions within a university community. Squirrel is a network of interactions among squirrels in a park. The Roman-Empire dataset represents a word co-occurrence network derived from the Roman Empire Wikipedia article. In this dataset, nodes correspond to individual words, and edges are formed between words that are either sequentially linked or syntactically related. Since the edges follow a specific order reflecting the sequence or syntactic structure of the text, the Roman-Empire dataset is inherently a directed graph. 

For graph classification tasks, we utilize the following datasets: NCI-1 and NCI-109 datasets involve classifying molecules as cancerous or non-cancerous. The node input features are represented as one-hot encodings of atom types, while edges signify chemical bonds. In datasets like Reddit-B, Reddit-5K, and Reddit-12K, interactions between users in Reddit discussion threads are captured. The primary task associated with these datasets is to determine the type of subreddit to which a discussion belongs. Collab comprises ego-networks from three distinct scientific collaboration fields. Unlike the previous datasets, Reddit tasks, and Collab, these datasets do not have node input features. Enzymes is a bioinformatics dataset for graph classification. It involves classifying enzymes based on their structures and functions. The BZR dataset is a small molecule dataset used for graph classification tasks. It is commonly employed for evaluating graph-based machine learning algorithms. MUTAG is another bioinformatics dataset for graph classification, primarily used for evaluating chemical informatics algorithms. The task is to predict mutagenicity. PTC is a bioinformatics dataset for graph classification, focusing on carcinogenicity prediction. The graphs represent chemical compounds. COX2 is a small molecule dataset, often used to assess graph-based machine learning models in chemistry-related tasks. The classification task is centered around predicting the inhibition of the COX-2 enzyme. Proteins is a bioinformatics dataset used for graph classification. The task is to classify proteins based on their functions. These datasets are from Tudataset \cite{mor_kri_bau-20a}.

In all these tasks, we intentionally avoid introducing structural input features such as node degrees or positional encodings. A summary of relevant dataset statistics is provided in Table \ref{tab:dataset} for reference.

\section{Conclusion and Future Direction}\label{sec:conclusion}
This survey has delved into the depths of over-squashing, unearthing its origins in information compression across distant nodes. The journey traversed a diverse array of strategies aimed at mitigating its impact – from innovative graph rewiring methods and curvature-based approaches to spectral techniques and the promise of graph transformers. As we tread this path, a nuanced interplay between over-smoothing and over-squashing has come into focus, demanding a balanced resolution. This exploration stands as a testament to the ongoing dialogue among researchers, driven by the pursuit of more refined and capable Graph Neural Networks. In closing, the quest to unravel over-squashing continues to be a beacon guiding our pursuit of more effective models, propelled by the dynamic nature of graph data.

While this survey highlights various techniques to mitigate the effects of over-squashing in GNNs, several promising directions remain open for further exploration:
\begin{enumerate}
\item Dynamic Graph Adaptation: Most existing models and rewiring techniques are designed for static graphs, where the node and edge structures remain fixed throughout the learning process. Although effective in certain scenarios, these methods often overlook the dynamic nature of real-world graphs. In static graphs, rewiring methods mitigate over-squashing by strategically adding edges or modifying the graph structure to improve long-range information flow. However, these fixed modifications may prove insufficient for dynamic graphs, where evolving connectivity patterns require adaptive solutions. To address this, Petrovi{'c} et al. \cite{pet-hua-pou_24a} proposed a Temporal Graph Rewiring (TGR) method specifically for dynamic graphs. Unlike static rewiring, TGR dynamically refines connectivity patterns by introducing time-aware edges that optimize information flow as the graph evolves. This adaptive strategy effectively reduces the risk of over-squashing, even in rapidly changing graph structures. While dynamic graphs offer flexibility in maintaining efficient information propagation, they also introduce practical challenges. Tracking and updating structural changes in real-time demands additional computational resources, making scalability a critical concern for dynamic graph rewiring strategies.

\item Handling Distributional Shifts: Graph representation learning often struggles to generalize effectively to out-of-distribution data, resulting in performance degradation in real-world scenarios. While certain techniques address distributional shifts in graph data, the interplay between these shifts and the over-squashing problem remains underexplored. Future research could investigate how mitigating over-squashing might enhance model robustness under such conditions.
\item Robustness to Noisy Data: Noise and adversarial perturbations in graph data can amplify the effects of over-squashing, further impairing model performance. Exploring techniques that improve GNN robustness — such as robust training methods or models inherently resilient to data perturbations — is crucial for enhancing performance in noisy environments.
\item Scalability and Efficiency: Existing over-squashing mitigation techniques often introduce additional computational overhead, posing challenges for large-scale graph datasets. Developing scalable algorithms or lightweight architectures that effectively mitigate over-squashing without excessive resource demands remains an important research direction.
\end{enumerate}

\section*{Acknowledgment}
I extend my heartfelt appreciation to Dr. Karmvir Singh Phogat for providing invaluable insights and essential feedback on the research problem explored in this article. His thoughtful comments significantly enriched the quality and lucidity of this study.

\bibliographystyle{./IEEEtran}
\bibliography{./IEEEabrv,bibfile_GNN_original}

\begin{thebibliography}{100}
\providecommand{\url}[1]{#1}
\csname url@samestyle\endcsname
\providecommand{\newblock}{\relax}
\providecommand{\bibinfo}[2]{#2}
\providecommand{\BIBentrySTDinterwordspacing}{\spaceskip=0pt\relax}
\providecommand{\BIBentryALTinterwordstretchfactor}{4}
\providecommand{\BIBentryALTinterwordspacing}{\spaceskip=\fontdimen2\font plus
\BIBentryALTinterwordstretchfactor\fontdimen3\font minus
  \fontdimen4\font\relax}
\providecommand{\BIBforeignlanguage}[2]{{%
\expandafter\ifx\csname l@#1\endcsname\relax
\typeout{** WARNING: IEEEtran.bst: No hyphenation pattern has been}%
\typeout{** loaded for the language `#1'. Using the pattern for}%
\typeout{** the default language instead.}%
\else
\language=\csname l@#1\endcsname
\fi
#2}}
\providecommand{\BIBdecl}{\relax}
\BIBdecl

\bibitem{ran_she_kout-15a}
S.~Ranshous, S.~Shen, D.~Koutra, S.~Harenberg, C.~Faloutsos, and N.~F.
  Samatova, ``Anomaly detection in dynamic networks: a survey,'' \emph{Wiley
  Interdisciplinary Reviews: Computational Statistics}, vol.~7, no.~3, pp.
  223--247, 2015.

\bibitem{les_mca-12a}
J.~Leskovec and J.~Mcauley, ``Learning to discover social circles in ego
  networks,'' \emph{Advances in neural information processing systems},
  vol.~25, 2012.

\bibitem{def_bre_van-16a}
M.~Defferrard, X.~Bresson, and P.~Vandergheynst, ``Convolutional neural
  networks on graphs with fast localized spectral filtering,'' \emph{Advances
  in neural information processing systems}, vol.~29, 2016.

\bibitem{gil_sch_ril-17a}
J.~Gilmer, S.~S. Schoenholz, P.~F. Riley, O.~Vinyals, and G.~E. Dahl, ``Neural
  message passing for quantum chemistry,'' in \emph{International conference on
  machine learning}.\hskip 1em plus 0.5em minus 0.4em\relax PMLR, 2017, pp.
  1263--1272.

\bibitem{ham_yin_les-17a}
W.~Hamilton, Z.~Ying, and J.~Leskovec, ``Inductive representation learning on
  large graphs,'' \emph{Advances in neural information processing systems},
  vol.~30, 2017.

\bibitem{che_li_bru-17a}
Z.~Chen, X.~Li, and J.~Bruna, ``Supervised community detection with line graph
  neural networks,'' \emph{arXiv preprint arXiv:1705.08415}, 2017.

\bibitem{min_gao_pen-21a}
S.~Min, Z.~Gao, J.~Peng, L.~Wang, K.~Qin, and B.~Fang, ``Stgsn—a
  spatial--temporal graph neural network framework for time-evolving social
  networks,'' \emph{Knowledge-Based Systems}, vol. 214, p. 106746, 2021.

\bibitem{wan_yuy-22a}
Y.~Wang, Y.~Zhao, Y.~Zhang, and T.~Derr, ``Collaboration-aware graph
  convolutional networks for recommendation systems,'' \emph{arXiv preprint
  arXiv:2207.06221}, 2022.

\bibitem{gao_wan-22a}
C.~Gao, X.~Wang, X.~He, and Y.~Li, ``Graph neural networks for recommender
  system,'' in \emph{Proceedings of the Fifteenth ACM International Conference
  on Web Search and Data Mining}, 2022, pp. 1623--1625.

\bibitem{chu_yao-22a}
Y.~Chu, J.~Yao, C.~Zhou, and H.~Yang, ``Graph neural networks in modern
  recommender systems,'' \emph{Graph Neural Networks: Foundations, Frontiers,
  and Applications}, pp. 423--445, 2022.

\bibitem{che_yeh_wan-22a}
H.~Chen, C.-C.~M. Yeh, F.~Wang, and H.~Yang, ``Graph neural transport networks
  with non-local attentions for recommender systems,'' in \emph{Proceedings of
  the ACM Web Conference 2022}, 2022, pp. 1955--1964.

\bibitem{yan_li-23a}
Y.~Yan, G.~Li \emph{et~al.}, ``Size generalizability of graph neural networks
  on biological data: Insights and practices from the spectral perspective,''
  \emph{arXiv preprint arXiv:2305.15611}, 2023.

\bibitem{jin_eis_son-21a}
B.~Jing, S.~Eismann, P.~N. Soni, and R.~O. Dror, ``Equivariant graph neural
  networks for 3d macromolecular structure,'' \emph{arXiv preprint
  arXiv:2106.03843}, 2021.

\bibitem{yua_yu_gui-22a}
H.~Yuan, H.~Yu, S.~Gui, and S.~Ji, ``Explainability in graph neural networks: A
  taxonomic survey,'' \emph{IEEE transactions on pattern analysis and machine
  intelligence}, vol.~45, no.~5, pp. 5782--5799, 2022.

\bibitem{zho_cui_hu-20a}
J.~Zhou, G.~Cui, S.~Hu, Z.~Zhang, C.~Yang, Z.~Liu, L.~Wang, C.~Li, and M.~Sun,
  ``Graph neural networks: A review of methods and applications,'' \emph{AI
  open}, vol.~1, pp. 57--81, 2020.

\bibitem{wu_pan_che-20a}
Z.~Wu, S.~Pan, F.~Chen, G.~Long, C.~Zhang, and S.~Y. Philip, ``A comprehensive
  survey on graph neural networks,'' \emph{IEEE transactions on neural networks
  and learning systems}, vol.~32, no.~1, pp. 4--24, 2020.

\bibitem{xu_hu_les-18a}
K.~Xu, W.~Hu, J.~Leskovec, and S.~Jegelka, ``How powerful are graph neural
  networks?'' \emph{arXiv preprint arXiv:1810.00826}, 2018.

\bibitem{sat-20a}
R.~Sato, ``A survey on the expressive power of graph neural networks,''
  \emph{arXiv preprint arXiv:2003.04078}, 2020.

\bibitem{xia_wan_dai-22a}
S.~Xiao, S.~Wang, Y.~Dai, and W.~Guo, ``Graph neural networks in node
  classification: survey and evaluation,'' \emph{Machine Vision and
  Applications}, vol.~33, pp. 1--19, 2022.

\bibitem{par_nev-2019a}
H.~Park and J.~Neville, ``Exploiting interaction links for node classification
  with deep graph neural networks.'' in \emph{IJCAI}, vol. 2019, 2019, pp.
  3223--3230.

\bibitem{wan_jin_zha-21a}
Y.~Wang, J.~Jin, W.~Zhang, Y.~Yu, Z.~Zhang, and D.~Wipf, ``Bag of tricks for
  node classification with graph neural networks,'' \emph{arXiv preprint
  arXiv:2103.13355}, 2021.

\bibitem{qiu_hua_xu-22a}
C.~Qiu, Z.~Huang, W.~Xu, and H.~Li, ``Vgaer: graph neural network
  reconstruction based community detection,'' \emph{arXiv preprint
  arXiv:2201.04066}, 2022.

\bibitem{wie_koh-20a}
O.~Wieder, S.~Kohlbacher, M.~Kuenemann, A.~Garon, P.~Ducrot, T.~Seidel, and
  T.~Langer, ``A compact review of molecular property prediction with graph
  neural networks,'' \emph{Drug Discovery Today: Technologies}, vol.~37, pp.
  1--12, 2020.

\bibitem{wan_liu_luo-22a}
Z.~Wang, M.~Liu, Y.~Luo, Z.~Xu, Y.~Xie, L.~Wang, L.~Cai, Q.~Qi, Z.~Yuan,
  T.~Yang \emph{et~al.}, ``Advanced graph and sequence neural networks for
  molecular property prediction and drug discovery,'' \emph{Bioinformatics},
  vol.~38, no.~9, pp. 2579--2586, 2022.

\bibitem{wu_che_wan-22a}
X.~Wu, Z.~Chen, W.~Wang, and A.~Jadbabaie, ``A non-asymptotic analysis of
  oversmoothing in graph neural networks,'' \emph{arXiv preprint
  arXiv:2212.10701}, 2022.

\bibitem{rus_bro_mis-23a}
T.~K. Rusch, M.~M. Bronstein, and S.~Mishra, ``A survey on oversmoothing in
  graph neural networks,'' \emph{arXiv preprint arXiv:2303.10993}, 2023.

\bibitem{luk_leh_fis-20a}
D.~Lukovnikov, J.~Lehmann, and A.~Fischer, ``Improving the long-range
  performance of gated graph neural networks,'' \emph{arXiv preprint
  arXiv:2007.09668}, 2020.

\bibitem{wu_jai_wri-21a}
Z.~Wu, P.~Jain, M.~Wright, A.~Mirhoseini, J.~E. Gonzalez, and I.~Stoica,
  ``Representing long-range context for graph neural networks with global
  attention,'' \emph{Advances in Neural Information Processing Systems},
  vol.~34, pp. 13\,266--13\,279, 2021.

\bibitem{mah_swe_kip-22a}
S.~Mahdavi, K.~Swersky, T.~Kipf, M.~Hashemi, C.~Thrampoulidis, and R.~Liao,
  ``Towards better out-of-distribution generalization of neural algorithmic
  reasoning tasks,'' \emph{arXiv preprint arXiv:2211.00692}, 2022.

\bibitem{aka-2023a}
S.~Akansha, ``Addressing the impact of localized training data in graph neural
  networks,'' \emph{arXiv preprint arXiv:2307.12689}, 2023.

\bibitem{bo_hu_wan-22a}
D.~Bo, B.~Hu, X.~Wang, Z.~Zhang, C.~Shi, and J.~Zhou, ``Regularizing graph
  neural networks via consistency-diversity graph augmentations,'' in
  \emph{Proceedings of the AAAI Conference on Artificial Intelligence},
  vol.~36, no.~4, 2022, pp. 3913--3921.

\bibitem{alo_yah-20a}
U.~Alon and E.~Yahav, ``On the bottleneck of graph neural networks and its
  practical implications,'' \emph{arXiv preprint arXiv:2006.05205}, 2020.

\bibitem{gir_mal_bou-22a}
J.~H. Giraldo, F.~D. Malliaros, and T.~Bouwmans, ``Understanding the
  relationship between over-smoothing and over-squashing in graph neural
  networks,'' \emph{arXiv preprint arXiv:2212.02374}, 2022.

\bibitem{gio_giu_far-23a}
F.~Di~Giovanni, L.~Giusti, F.~Barbero, G.~Luise, P.~Lio, and M.~M. Bronstein,
  ``On over-squashing in message passing neural networks: The impact of width,
  depth, and topology,'' in \emph{International Conference on Machine
  Learning}.\hskip 1em plus 0.5em minus 0.4em\relax PMLR, 2023, pp. 7865--7885.

\bibitem{top_gio_cha-21a}
J.~Topping, F.~Di~Giovanni, B.~P. Chamberlain, X.~Dong, and M.~M. Bronstein,
  ``Understanding over-squashing and bottlenecks on graphs via curvature,''
  \emph{arXiv preprint arXiv:2111.14522}, 2021.

\bibitem{att-bus_24a}
H.~Attali, D.~Buscaldi, and N.~Pernelle, ``Rewiring techniques to mitigate
  oversquashing and oversmoothing in gnns: A survey,'' \emph{arXiv preprint
  arXiv:2411.17429}, 2024.

\bibitem{yad_24a}
N.~Yadati, ``Oversquashing in hypergraph neural networks,'' in \emph{The Third
  Learning on Graphs Conference}.

\bibitem{zha_yao-22a}
Y.~Zhang and Q.~Yao, ``Knowledge graph reasoning with relational digraph,'' in
  \emph{Proceedings of the ACM Web Conference 2022}, 2022, pp. 912--924.

\bibitem{bla_wan_nay-23a}
M.~Black, Z.~Wan, A.~Nayyeri, and Y.~Wang, ``Understanding oversquashing in
  gnns through the lens of effective resistance,'' in \emph{International
  Conference on Machine Learning}.\hskip 1em plus 0.5em minus 0.4em\relax PMLR,
  2023, pp. 2528--2547.

\bibitem{cha-rag_89a}
A.~K. Chandra, P.~Raghavan, W.~L. Ruzzo, and R.~Smolensky, ``The electrical
  resistance of a graph captures its commute and cover times,'' in
  \emph{Proceedings of the twenty-first annual ACM symposium on Theory of
  computing}, 1989, pp. 574--586.

\bibitem{gio_rus_bro-23a}
F.~Di~Giovanni, T.~K. Rusch, M.~M. Bronstein, A.~Deac, M.~Lackenby, S.~Mishra,
  and P.~Veli{\v{c}}kovi{\'c}, ``How does over-squashing affect the power of
  gnns?'' \emph{arXiv preprint arXiv:2306.03589}, 2023.

\bibitem{ngu_hie-23a}
K.~Nguyen, N.~M. Hieu, V.~D. Nguyen, N.~Ho, S.~Osher, and T.~M. Nguyen,
  ``Revisiting over-smoothing and over-squashing using ollivier-ricci
  curvature,'' in \emph{International Conference on Machine Learning}.\hskip
  1em plus 0.5em minus 0.4em\relax PMLR, 2023, pp. 25\,956--25\,979.

\bibitem{yin_you_mor-18a}
Z.~Ying, J.~You, C.~Morris, X.~Ren, W.~Hamilton, and J.~Leskovec,
  ``Hierarchical graph representation learning with differentiable pooling,''
  \emph{Advances in neural information processing systems}, vol.~31, 2018.

\bibitem{luz_day_lio-19a}
E.~Luzhnica, B.~Day, and P.~Lio, ``Clique pooling for graph classification,''
  \emph{arXiv preprint arXiv:1904.00374}, 2019.

\bibitem{san_rot_lie-23a}
C.~Sanders, A.~Roth, and T.~Liebig, ``Curvature-based pooling within graph
  neural networks,'' \emph{arXiv preprint arXiv:2308.16516}, 2023.

\bibitem{kar_ban-22a}
K.~Karhadkar, P.~K. Banerjee, and G.~Mont{\'u}far, ``Fosr: First-order spectral
  rewiring for addressing oversquashing in gnns,'' \emph{arXiv preprint
  arXiv:2210.11790}, 2022.

\bibitem{liu_zho_pan-23a}
Y.~Liu, C.~Zhou, S.~Pan, J.~Wu, Z.~Li, H.~Chen, and P.~Zhang, ``Curvdrop: A
  ricci curvature based approach to prevent graph neural networks from
  over-smoothing and over-squashing,'' in \emph{Proceedings of the ACM Web
  Conference 2023}, 2023, pp. 221--230.

\bibitem{ban_kar_wan-22a}
P.~K. Banerjee, K.~Karhadkar, Y.~G. Wang, U.~Alon, and G.~Mont{\'u}far,
  ``Oversquashing in gnns through the lens of information contraction and graph
  expansion,'' in \emph{2022 58th Annual Allerton Conference on Communication,
  Control, and Computing (Allerton)}.\hskip 1em plus 0.5em minus 0.4em\relax
  IEEE, 2022, pp. 1--8.

\bibitem{dea_lac_vel-22a}
A.~Deac, M.~Lackenby, and P.~Veli{\v{c}}kovi{\'c}, ``Expander graph
  propagation,'' in \emph{Learning on Graphs Conference}.\hskip 1em plus 0.5em
  minus 0.4em\relax PMLR, 2022, pp. 38--1.

\bibitem{cai-tru_23a}
C.~Cai, T.~S. Hy, R.~Yu, and Y.~Wang, ``On the connection between mpnn and
  graph transformer,'' \emph{arXiv preprint arXiv:2301.11956}, 2023.

\bibitem{gas_etal-19b}
J.~Gasteiger, S.~Wei{\ss}enberger, and S.~G{\"u}nnemann, ``Diffusion improves
  graph learning,'' \emph{Advances in neural information processing systems},
  vol.~32, 2019.

\bibitem{wan-yin_20a}
G.~Wang, R.~Ying, J.~Huang, and J.~Leskovec, ``Multi-hop attention graph neural
  network,'' \emph{arXiv preprint arXiv:2009.14332}, 2020.

\bibitem{li-li-zha_24a}
H.~Li, C.~Li, J.~Zhang, Y.~Ouyang, and W.~Rong, ``Addressing over-squashing in
  gnns with graph rewiring and ordered neurons,'' in \emph{2024 International
  Joint Conference on Neural Networks (IJCNN)}.\hskip 1em plus 0.5em minus
  0.4em\relax IEEE, 2024, pp. 1--8.

\bibitem{lin-lu-yan_11a}
Y.~Lin, L.~Lu, and S.-T. Yau, ``Ricci curvature of graphs,'' \emph{Tohoku
  Mathematical Journal, Second Series}, vol.~63, no.~4, pp. 605--627, 2011.

\bibitem{gab_yur_sol-22a}
R.~B. Gabrielsson, M.~Yurochkin, and J.~Solomon, ``Rewiring with positional
  encodings for gnns,'' 2022.

\bibitem{ram-gal_22a}
L.~Ramp{\'a}{\v{s}}ek, M.~Galkin, V.~P. Dwivedi, A.~T. Luu, G.~Wolf, and
  D.~Beaini, ``Recipe for a general, powerful, scalable graph transformer,''
  \emph{Advances in Neural Information Processing Systems}, vol.~35, pp.
  14\,501--14\,515, 2022.

\bibitem{att-bus-per_25a}
H.~Attali, D.~Buscaldi, and N.~Pernelle, ``Curvature constrained mpnns:
  Improving message passing with local structural properties,'' \emph{Data \&
  Knowledge Engineering}, vol. 156, p. 102382, 2025.

\bibitem{qia-man_23a}
C.~Qian, A.~Manolache, K.~Ahmed, Z.~Zeng, G.~V.~d. Broeck, M.~Niepert, and
  C.~Morris, ``Probabilistically rewired message-passing neural networks,''
  \emph{arXiv preprint arXiv:2310.02156}, 2023.

\bibitem{ahm-zen_22a}
K.~Ahmed, Z.~Zeng, M.~Niepert, and G.~V.~d. Broeck, ``Simple: A gradient
  estimator for $ k $-subset sampling,'' \emph{arXiv preprint
  arXiv:2210.01941}, 2022.

\bibitem{gut_don_bro-23a}
B.~Gutteridge, X.~Dong, M.~M. Bronstein, and F.~Di~Giovanni, ``Drew:
  Dynamically rewired message passing with delay,'' in \emph{International
  Conference on Machine Learning}.\hskip 1em plus 0.5em minus 0.4em\relax PMLR,
  2023, pp. 12\,252--12\,267.

\bibitem{abb-rad_22a}
R.~Abboud, R.~Dimitrov, and I.~I. Ceylan, ``Shortest path networks for graph
  property prediction,'' in \emph{Learning on Graphs Conference}.\hskip 1em
  plus 0.5em minus 0.4em\relax PMLR, 2022, pp. 5--1.

\bibitem{tor_mic-22a}
D.~Tortorella and A.~Micheli, ``Leave graphs alone: Addressing over-squashing
  without rewiring,'' \emph{arXiv preprint arXiv:2212.06538}, 2022.

\bibitem{fes-web_24a}
L.~Fesser and M.~Weber, ``Mitigating over-smoothing and over-squashing using
  augmentations of forman-ricci curvature,'' in \emph{Learning on Graphs
  Conference}.\hskip 1em plus 0.5em minus 0.4em\relax PMLR, 2024, pp. 19--1.

\bibitem{arn_and_beg-22a}
A.~Arnaiz-Rodr{\'\i}guez, A.~Begga, F.~Escolano, and N.~Oliver, ``Diffwire:
  Inductive graph rewiring via the lov$\backslash$'asz bound,'' \emph{arXiv
  preprint arXiv:2206.07369}, 2022.

\bibitem{fed-ame-ame_24a}
\BIBentryALTinterwordspacing
F.~Barbero, A.~Velingker, A.~Saberi, M.~M. Bronstein, and F.~D. Giovanni,
  ``Locality-aware graph rewiring in {GNN}s,'' in \emph{The Twelfth
  International Conference on Learning Representations}, 2024. [Online].
  Available: \url{https://openreview.net/forum?id=4Ua4hKiAJX}
\BIBentrySTDinterwordspacing

\bibitem{bea_pas_vin-21a}
D.~Beaini, S.~Passaro, V.~L{\'e}tourneau, W.~Hamilton, G.~Corso, and
  P.~Li{\`o}, ``Directional graph networks,'' in \emph{International Conference
  on Machine Learning}.\hskip 1em plus 0.5em minus 0.4em\relax PMLR, 2021, pp.
  748--758.

\bibitem{sha_shi_and-23a}
Z.~Shao, D.~Shi, A.~Han, Y.~Guo, Q.~Zhao, and J.~Gao, ``Unifying over-smoothing
  and over-squashing in graph neural networks: A physics informed approach and
  beyond,'' \emph{arXiv preprint arXiv:2309.02769}, 2023.

\bibitem{gra_bac_gal-22a}
A.~Gravina, D.~Bacciu, and C.~Gallicchio, ``Anti-symmetric dgn: A stable
  architecture for deep graph networks,'' \emph{arXiv preprint
  arXiv:2210.09789}, 2022.

\bibitem{sun_li_yua-22a}
Q.~Sun, J.~Li, H.~Yuan, X.~Fu, H.~Peng, C.~Ji, Q.~Li, and P.~S. Yu,
  ``Position-aware structure learning for graph topology-imbalance by relieving
  under-reaching and over-squashing,'' in \emph{Proceedings of the 31st ACM
  International Conference on Information \& Knowledge Management}, 2022, pp.
  1848--1857.

\bibitem{she-qin-zha_24a}
D.~Shen, C.~Qin, Q.~Zhang, H.~Zhu, and H.~Xiong, ``Handling over-smoothing and
  over-squashing in graph convolution with maximization operation,'' \emph{IEEE
  Transactions on Neural Networks and Learning Systems}, 2024.

\bibitem{che_zha_li-22a}
R.~Chen, S.~Zhang, Y.~Li \emph{et~al.}, ``Redundancy-free message passing for
  graph neural networks,'' \emph{Advances in Neural Information Processing
  Systems}, vol.~35, pp. 4316--4327, 2022.

\bibitem{chu-97a}
F.~R. Chung, \emph{Spectral graph theory}.\hskip 1em plus 0.5em minus
  0.4em\relax American Mathematical Soc., 1997, vol.~92.

\bibitem{sel_65a}
A.~Selberg, ``On the estimation of fourier coefficients of modular forms,'' in
  \emph{Proceedings of Symposia in Pure Mathematics}.\hskip 1em plus 0.5em
  minus 0.4em\relax American Mathematical Society, 1965, pp. 1--15.

\bibitem{buh-hei_09a}
T.~B{\"u}hler and M.~Hein, ``Spectral clustering based on the graph
  p-laplacian,'' in \emph{Proceedings of the 26th annual international
  conference on machine learning}, 2009, pp. 81--88.

\bibitem{all-bha-lat_16a}
Z.~Allen-Zhu, A.~Bhaskara, S.~Lattanzi, V.~Mirrokni, and L.~Orecchia,
  ``Expanders via local edge flips,'' in \emph{Proceedings of the
  twenty-seventh annual ACM-SIAM symposium on Discrete algorithms}.\hskip 1em
  plus 0.5em minus 0.4em\relax SIAM, 2016, pp. 259--269.

\bibitem{fed-gue-ada_06a}
T.~Feder, A.~Guetz, M.~Mihail, and A.~Saberi, ``A local switch markov chain on
  given degree graphs with application in connectivity of peer-to-peer
  networks,'' in \emph{2006 47th Annual IEEE Symposium on Foundations of
  Computer Science (FOCS'06)}.\hskip 1em plus 0.5em minus 0.4em\relax IEEE,
  2006, pp. 69--76.

\bibitem{coo-dye-gre_19a}
C.~Cooper, M.~Dyer, C.~Greenhill, and A.~Handley, ``The flip markov chain for
  connected regular graphs,'' \emph{Discrete Applied Mathematics}, vol. 254,
  pp. 56--79, 2019.

\bibitem{kre_bea_ham-21a}
D.~Kreuzer, D.~Beaini, W.~Hamilton, V.~L{\'e}tourneau, and P.~Tossou,
  ``Rethinking graph transformers with spectral attention,'' \emph{Advances in
  Neural Information Processing Systems}, vol.~34, pp. 21\,618--21\,629, 2021.

\bibitem{oon_suz-19a}
K.~Oono and T.~Suzuki, ``Graph neural networks exponentially lose expressive
  power for node classification,'' \emph{arXiv preprint arXiv:1905.10947},
  2019.

\bibitem{tor-hol_24a}
F.~Tori, V.~Holst, and V.~Ginis, ``The effectiveness of curvature-based
  rewiring and the role of hyperparameters in gnns revisited,'' \emph{arXiv
  preprint arXiv:2407.09381}, 2024.

\bibitem{tor_mic-23a}
D.~Tortorella and A.~Micheli, ``Is rewiring actually helpful in graph neural
  networks?'' \emph{arXiv preprint arXiv:2305.19717}, 2023.

\bibitem{ton-mar_24a}
\BIBentryALTinterwordspacing
J.~T{\"o}nshoff, M.~Ritzert, E.~Rosenbluth, and M.~Grohe, ``Where did the gap
  go? reassessing the long-range graph benchmark,'' \emph{Transactions on
  Machine Learning Research}, 2024. [Online]. Available:
  \url{https://openreview.net/forum?id=Nm0WX86sKv}
\BIBentrySTDinterwordspacing

\bibitem{li_han__wu-18a}
Q.~Li, Z.~Han, and X.-M. Wu, ``Deeper insights into graph convolutional
  networks for semi-supervised learning,'' in \emph{Proceedings of the AAAI
  conference on artificial intelligence}, vol.~32, no.~1, 2018.

\bibitem{hoa-mae-mur_21a}
N.~Hoang, T.~Maehara, and T.~Murata, ``Revisiting graph neural networks: Graph
  filtering perspective,'' in \emph{2020 25th International Conference on
  Pattern Recognition (ICPR)}.\hskip 1em plus 0.5em minus 0.4em\relax IEEE,
  2021, pp. 8376--8383.

\bibitem{rus-cha_22a}
T.~K. Rusch, B.~Chamberlain, J.~Rowbottom, S.~Mishra, and M.~Bronstein,
  ``Graph-coupled oscillator networks,'' in \emph{International Conference on
  Machine Learning}.\hskip 1em plus 0.5em minus 0.4em\relax PMLR, 2022, pp.
  18\,888--18\,909.

\bibitem{shi_guo_zhi-23a}
D.~Shi, Y.~Guo, Z.~Shao, and J.~Gao, ``How curvature enhance the adaptation
  power of framelet gcns,'' \emph{arXiv preprint arXiv:2307.09768}, 2023.

\bibitem{wu-sou-zha_19a}
F.~Wu, A.~Souza, T.~Zhang, C.~Fifty, T.~Yu, and K.~Weinberger, ``Simplifying
  graph convolutional networks,'' in \emph{International conference on machine
  learning}.\hskip 1em plus 0.5em minus 0.4em\relax Pmlr, 2019, pp. 6861--6871.

\bibitem{zho-hua-li_20a}
K.~Zhou, X.~Huang, Y.~Li, D.~Zha, R.~Chen, and X.~Hu, ``Towards deeper graph
  neural networks with differentiable group normalization,'' \emph{Advances in
  neural information processing systems}, vol.~33, pp. 4917--4928, 2020.

\bibitem{gas_etal-19a}
J.~Gasteiger, A.~Bojchevski, and S.~G{\"u}nnemann, ``Predict then propagate:
  Graph neural networks meet personalized pagerank,'' in \emph{International
  Conference on Learning Representations (ICLR)}, 2019.

\bibitem{man-isl_25a}
K.~Mancini and I.~Rekik, ``Duognn: Topology-aware graph neural network with
  homophily and heterophily interaction-decoupling,'' in \emph{Graphs in
  Biomedical Image Analysis}, S.-A. Ahmadi and A.~Kazi, Eds.\hskip 1em plus
  0.5em minus 0.4em\relax Cham: Springer Nature Switzerland, 2025, pp.
  129--140.

\bibitem{man-rek_25a}
------, ``Deltagnn: Graph neural network with information flow control,''
  \emph{arXiv preprint arXiv:2501.06002}, 2025.

\bibitem{yun_jeo_kim-19a}
S.~Yun, M.~Jeong, R.~Kim, J.~Kang, and H.~J. Kim, ``Graph transformer
  networks,'' \emph{Advances in neural information processing systems},
  vol.~32, 2019.

\bibitem{che_wu_dai-22a}
C.~Chen, Y.~Wu, Q.~Dai, H.-Y. Zhou, M.~Xu, S.~Yang, X.~Han, and Y.~Yu, ``A
  survey on graph neural networks and graph transformers in computer vision: a
  task-oriented perspective,'' \emph{arXiv preprint arXiv:2209.13232}, 2022.

\bibitem{cai_lam-20a}
D.~Cai and W.~Lam, ``Graph transformer for graph-to-sequence learning,'' in
  \emph{Proceedings of the AAAI conference on artificial intelligence},
  vol.~34, no.~05, 2020, pp. 7464--7471.

\bibitem{yin_cai_luo-21a}
C.~Ying, T.~Cai, S.~Luo, S.~Zheng, G.~Ke, D.~He, Y.~Shen, and T.-Y. Liu, ``Do
  transformers really perform badly for graph representation?'' \emph{Advances
  in Neural Information Processing Systems}, vol.~34, pp. 28\,877--28\,888,
  2021.

\bibitem{he_hoo_lau-23a}
X.~He, B.~Hooi, T.~Laurent, A.~Perold, Y.~LeCun, and X.~Bresson, ``A
  generalization of vit/mlp-mixer to graphs,'' in \emph{International
  Conference on Machine Learning}.\hskip 1em plus 0.5em minus 0.4em\relax PMLR,
  2023, pp. 12\,724--12\,745.

\bibitem{oll_09a}
Y.~Ollivier, ``Ricci curvature of markov chains on metric spaces,''
  \emph{Journal of Functional Analysis}, vol. 256, no.~3, pp. 810--864, 2009.

\bibitem{nak_18a}
K.~Nakajima, ``Reservoir computing: Theory, physical implementations, and
  applications,'' \emph{IEICE Technical Report; IEICE Tech. Rep.}, vol. 118,
  no. 220, pp. 149--154, 2018.

\bibitem{pei_wei_cha-20a}
H.~Pei, B.~Wei, K.~C.-C. Chang, Y.~Lei, and B.~Yang, ``Geom-gcn: Geometric
  graph convolutional networks,'' \emph{arXiv preprint arXiv:2002.05287}, 2020.

\bibitem{mcc_nig_ren-00a}
A.~K. McCallum, K.~Nigam, J.~Rennie, and K.~Seymore, ``Automating the
  construction of internet portals with machine learning,'' \emph{Information
  Retrieval}, vol.~3, pp. 127--163, 2000.

\bibitem{sen_nam-08a}
P.~Sen, G.~Namata, M.~Bilgic, L.~Getoor, B.~Galligher, and T.~Eliassi-Rad,
  ``Collective classification in network data,'' \emph{AI magazine}, vol.~29,
  no.~3, pp. 93--93, 2008.

\bibitem{nam_lon-12a}
G.~Namata, B.~London, L.~Getoor, B.~Huang, and U.~Edu, ``Query-driven active
  surveying for collective classification,'' in \emph{10th international
  workshop on mining and learning with graphs}, vol.~8, 2012, p.~1.

\bibitem{tan_sun_wan-09a}
J.~Tang, J.~Sun, C.~Wang, and Z.~Yang, ``Social influence analysis in
  large-scale networks,'' in \emph{Proceedings of the 15th ACM SIGKDD
  international conference on Knowledge discovery and data mining}, 2009, pp.
  807--816.

\bibitem{roz_all_sar-21a}
B.~Rozemberczki, C.~Allen, and R.~Sarkar, ``Multi-scale attributed node
  embedding,'' \emph{Journal of Complex Networks}, vol.~9, no.~2, p. cnab014,
  2021.

\bibitem{pla-kuz_23a}
\BIBentryALTinterwordspacing
O.~Platonov, D.~Kuznedelev, M.~Diskin, A.~Babenko, and L.~Prokhorenkova, ``A
  critical look at the evaluation of {GNN}s under heterophily: Are we really
  making progress?'' in \emph{The Eleventh International Conference on Learning
  Representations}, 2023. [Online]. Available:
  \url{https://openreview.net/forum?id=tJbbQfw-5wv}
\BIBentrySTDinterwordspacing

\bibitem{mor_kri_bau-20a}
C.~Morris, N.~M. Kriege, F.~Bause, K.~Kersting, P.~Mutzel, and M.~Neumann,
  ``Tudataset: A collection of benchmark datasets for learning with graphs,''
  \emph{arXiv preprint arXiv:2007.08663}, 2020.

\bibitem{pet-hua-pou_24a}
K.~Petrovi{\'c}, S.~Huang, F.~Poursafaei, and P.~Veli{\v{c}}kovi{\'c},
  ``Temporal graph rewiring with expander graphs,'' \emph{arXiv preprint
  arXiv:2406.02362}, 2024.

\end{thebibliography}
\end{document}